%% file: main.tex
\documentclass[10pt,twocolumn,letterpaper]{article}

\usepackage[pagenumbers]{iccv} 


\definecolor{iccvblue}{rgb}{0.21,0.49,0.74}
\usepackage[pagebackref,breaklinks,colorlinks,allcolors=iccvblue]{hyperref}

\usepackage[capitalize]{cleveref}
\crefname{section}{Sec.}{Secs.}
\Crefname{section}{Section}{Sections}
\Crefname{table}{Table}{Tables}
\crefname{table}{Tab.}{Tabs.}
\usepackage{bm}
\usepackage{capt-of}
\usepackage{cuted}
\usepackage{multirow}
\usepackage{relsize}
\usepackage{amsmath}
\usepackage{xcolor}
\usepackage{colortbl, booktabs}
\usepackage{caption}
\usepackage{subcaption}
\usepackage{amssymb}
\usepackage{bbding}
\usepackage[accsupp]{axessibility}

\usepackage[normalem]{ulem}

\def\paperID{8319} 
\def\confName{ICCV}
\def\confYear{2025}

\title{IMG: Calibrating Diffusion Models via Implicit Multimodal Guidance}

\author{%
Jiayi Guo$^{1,2}$\thanks{Equal contribution. $^\dagger$Corresponding authors.}\ \ \ 
Chuanhao Yan$^{1,2}$\footnotemark[1]\ \ \ 
Xingqian Xu$^1$\ \ \ 
Yulin Wang$^2$\ \ \ 
Kai Wang$^1$\\
Gao Huang$^{2\dagger}$ \ \ \ 
Humphrey Shi$^{1\dagger}$\\
{\small$^1$SHI Labs @ Georgia Tech \ \ \
$^2$Tsinghua University} \\ 
{\small \textbf{\href{https://github.com/SHI-Labs/IMG-Multimodal-Diffusion-Alignment}{\texttt{\textcolor{magenta}{https://github.com/SHI-Labs/IMG-Multimodal-Diffusion-Alignment}}}}}
\vspace{-30mm}
}

\begin{document}
\maketitle

\begin{strip} 
    \centering
  \includegraphics[width=1.\linewidth]{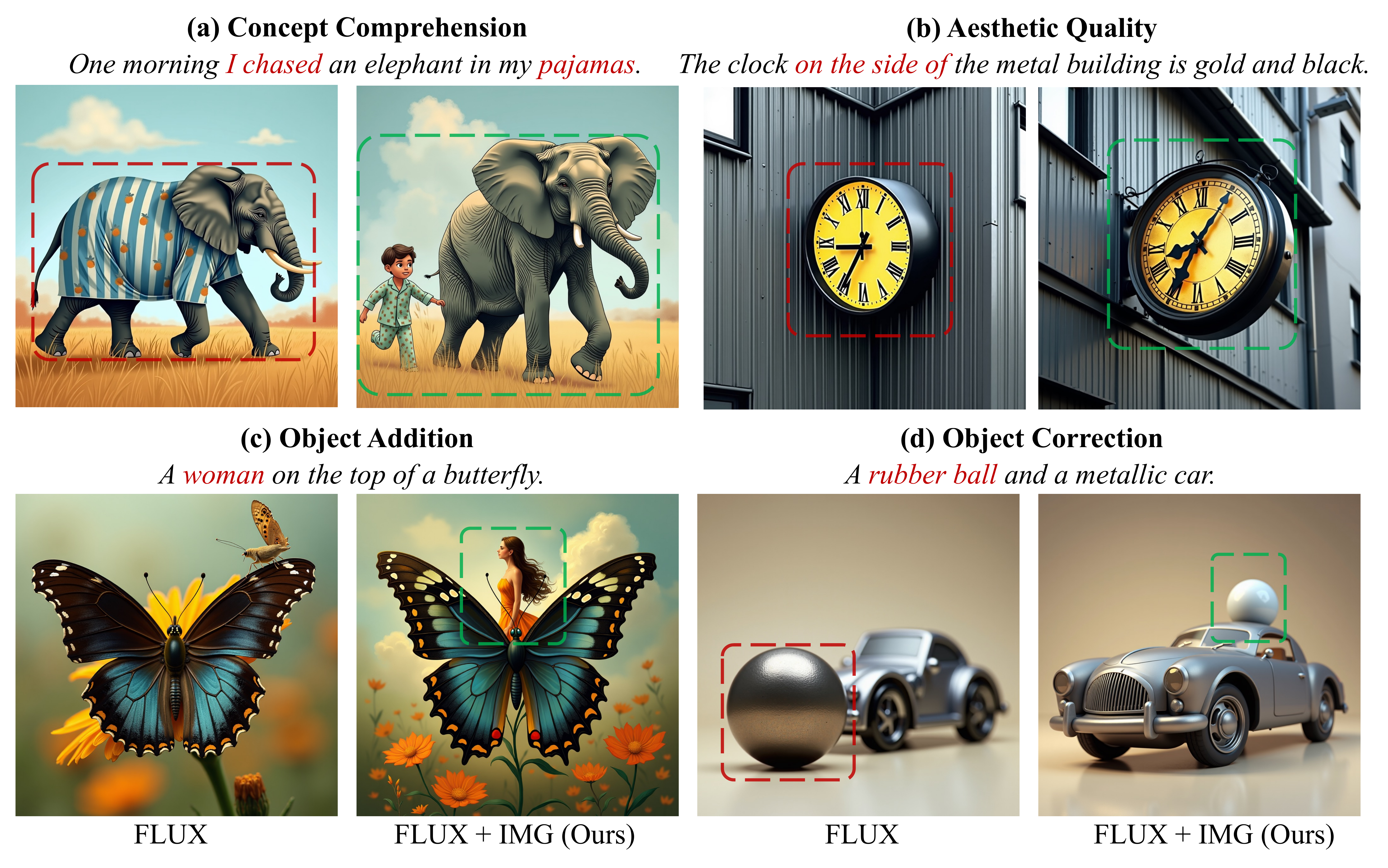}
    \vspace{-8mm}   
  \captionof{figure}{
\textbf{The multimodal misalignment issue.} Even the latest state-of-the-art diffusion model, FLUX.1 [dev] (FLUX)~\cite{flux}, may overlook or misinterpret concepts in prompts. Assisted with our proposed Implicit Multimodal Guidance (IMG) framework, the prompt-image misalignment issues are significantly mitigated in various aspects such as concept comprehension, aesthetic quality, object addition, and correction. In each case, both images are generated with the same random seed for fair comparison.}
  \label{fig:teaser}
\end{strip}

\maketitle
\input{sec/0_abstract}

\input{sec/1_intro}

\input{sec/2_related}
\input{sec/3_1_method}

\input{sec/3_2_method}
\input{sec/3_3_method}
\input{sec/3_4_method}
\input{sec/4_exp}
\input{sec/5_conclusion}

{
    \small
    \bibliographystyle{ieeenat_fullname}
    \bibliography{main}
}
\clearpage
\clearpage
\input{sec/appendix}

\end{document}

%% file: sec/0_abstract.tex
\begin{abstract}
Ensuring precise multimodal alignment between diffusion-generated images and input prompts has been a long-standing challenge. Earlier works finetune diffusion weight using high-quality preference data, which tends to be limited and difficult to scale up. Recent editing-based methods further refine local regions of generated images but may compromise overall image quality. In this work, we propose \textbf{I}mplicit \textbf{M}ultimodal \textbf{G}uidance (\textbf{IMG}), a novel re-generation-based multimodal alignment framework that requires no extra data or editing operations. Specifically, given a generated image and its prompt, IMG a) utilizes a multimodal large language model (MLLM) to identify misalignments; b) introduces an Implicit Aligner that manipulates diffusion conditioning features to reduce misalignments and enable re-generation; and c) formulates the re-alignment goal into a trainable objective, namely Iteratively Updated Preference Objective. Extensive qualitative and quantitative evaluations on SDXL, SDXL-DPO, and FLUX show that IMG outperforms existing alignment methods. Furthermore, IMG acts as a flexible plug-and-play adapter, seamlessly enhancing prior finetuning-based alignment methods. Our code will be available at \href{https://github.com/SHI-Labs/IMG-Multimodal-Diffusion-Alignment}{\textit{https://github.com/SHI-Labs/IMG--Multimodal-Diffusion-Alignment}}.

\end{abstract}

%% file: sec/1_intro.tex
\section{Introduction}
\label{sec:intro}

Recently, diffusion models have become powerful text-to-image (T2I) generation tools capable of producing diverse and realistic images. However, most methods still face input-output misalignment challenges based on human inspection. More specifically, these models may overlook or misinterpret a few aspects of prompts, thus creating undesired visual results with misinterpretations.~\cref{fig:teaser} shows evidence that even the latest state-of-the-art diffusion model, FLUX~\cite{flux}, generates images that require further refinement in terms of prompt awareness and adherence.

\begin{figure}[t]
    \centering
    \includegraphics[width=0.95\linewidth]{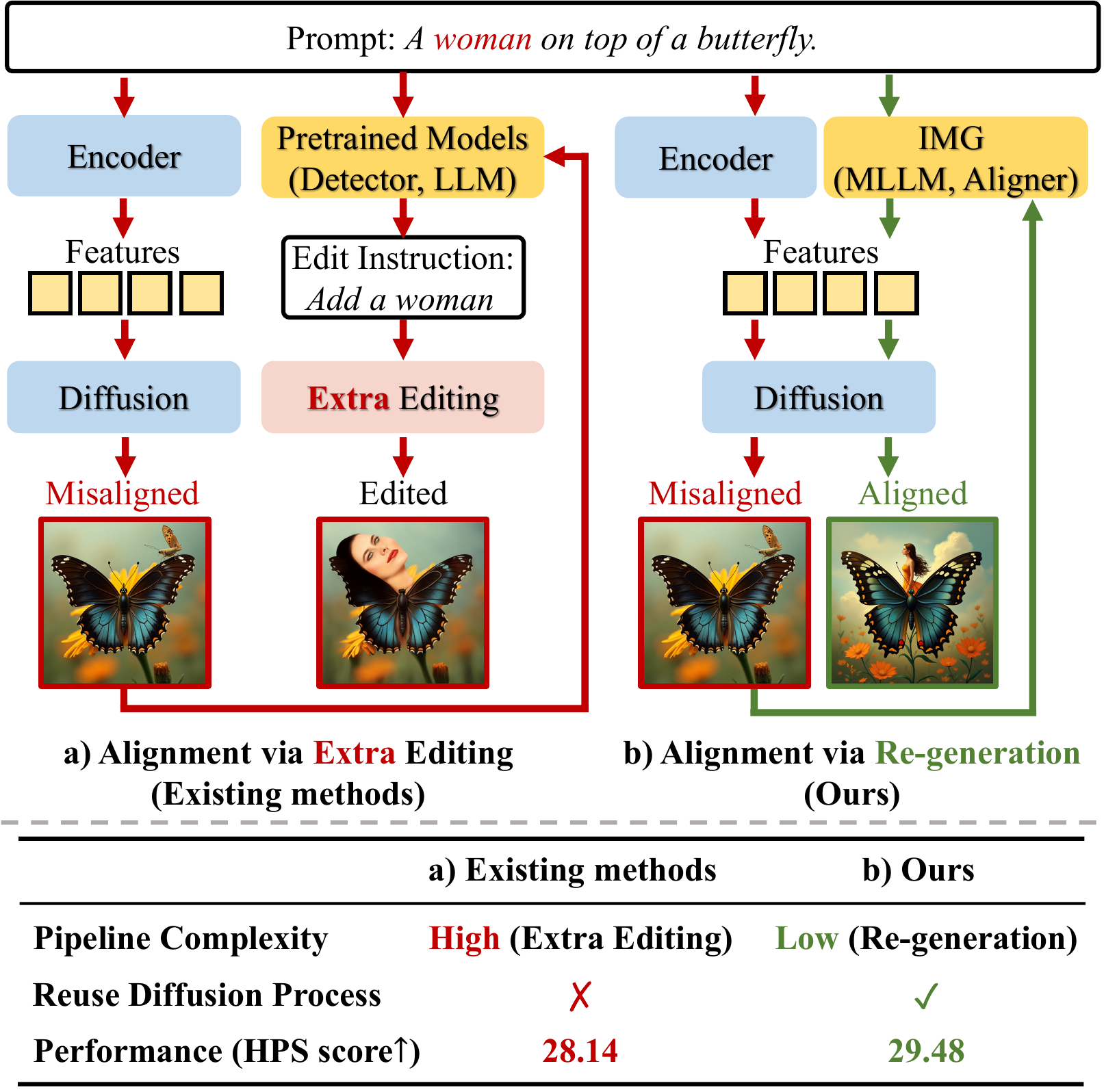}
      \vspace{-2mm}   
    \caption{\textbf{Comparison between our Implicit Multimodal Guidance (IMG) and existing editing-based alignment methods.} a) Existing methods require additional editing operations that improve alignment in local regions but may compromise overall image quality. b) In contrast, IMG employs a re-generation-based alignment framework by manipulating diffusion conditioning features, ensuring pipeline simplicity and high-quality outputs.}
    \label{fig:illu1}
    \vspace{-3mm}
\end{figure}

Due to the aforementioned challenges, exploring methods that improve prompt-image alignment has become an emerging and critical area of research. Early works in this field primarily employ the \textit{preference-based weight finetuning}. Methods such as~\cite{sdxl,emu} finetune diffusion models on high-quality prompt-image pairs. Later studies~\cite{dpok,directlyft,dpo,diffusionrl,spin} apply reinforcement learning from human feedback (RLHF), exploring rewarding algorithms and datasets that are proven to be effective for large language models (LLMs)~\cite{gpt4,llama3}. However, these methods are constrained by the limited availability of high-quality finetuning data, which is difficult to scale up further.
In contrast, the recent \textit{LLM-driven image editing} method~\cite{sld} eliminates the need for fine-tuning model weights. It combines an open-set detector~\cite{ovvit} with an LLM~\cite{gpt4} to identify misalignments in generated images and produces language instructions for image editing (see~\cref{fig:illu1}a). Leveraging powerful LLMs for verification and reflection on generation results is a promising research direction~\cite{dsr1,oaio1,oaio3}. However, the current editing pipeline primarily focuses on improving alignment in locally edited regions, and we empirically find that the overall image quality often fails to maintain the pre-editing level. Additionally, its detector sometimes omits critical misalignments, resulting in inaccurate editing instructions from the LLM.

In this work, we explore enhancing alignment performance without additional data required by finetuning-based methods or extra editing operations in editing-based methods. To this end, we propose the \textit{Implicit Multimodal Guidance} (IMG), a novel diffusion alignment framework that improves prompt adherence through a simple yet effective re-generation process. Specifically, IMG involves a multimodal large language model (MLLM) and a newly introduced Implicit Aligner. The MLLM first detects misalignment between the prompts and the images generated by diffusion models. Subsequently, the Implicit Aligner takes as input both features of MLLM and misaligned images, producing better-aligned diffusion conditioning features to reduce misalignment and enable re-generation. To train the Implicit Aligner, we introduce an Iteratively Updated Preference Objective, which combines direct preference optimization~\cite{dpollm} and self-play finetuning~\cite{spinllm}.

To conclude, IMG distinguishes itself from prior approaches by: a) serving as a flexible plug-and-play adapter that seamlessly integrates with both base diffusion models and their finetuned versions once trained, b) leveraging the native diffusion generation process to maintain pipeline simplicity and high-quality outputs, and c) eliminating the need for additional editing operations, making it a more straightforward and efficient solution. Built on these principles, we conduct extensive qualitative and quantitative evaluations across popular T2I models, including SDXL~\cite{sdxl}, SDXL-DPO~\cite{dpo}, and FLUX~\cite{flux}, as well as both finetuning-based and editing-based alignment methods. Results consistently demonstrate that IMG effectively re-aligns image outputs to improved versions and outperforms existing alignment methods.

%% file: sec/2_related.tex
\section{Related Work}
\label{sec:related}

\vspace{1mm}
\noindent\textbf{Diffusion model alignment} is a research area focusing on improving the prompt adherence of diffusion models in terms of human perception.
Early efforts~\cite{sd,emu,sdxl,hps,xu2022versatile} directly finetune diffusion models on high-quality datasets to produce visually appealing results. Later studies conduct rewarding algorithms based on reinforcement learning~\cite{diffusionrl,directlyft,dpok,ImageReward,social} and preference learning~\cite{dpo,d3po,spin} using datasets labeled with human preferences~\cite{pick,risa}. 
For example, DDPO~\cite{diffusionrl} utilizes black-box reward functions to optimize diffusion models within specific prompt sets.
Diffusion-DPO~\cite{dpo} re-formulates the direct preference optimization~\cite{dpollm} into a differentiable diffusion objective via the evidence lower bound.
Inspired by the rapid advances of large language models (LLMs)~\cite{llama3,llava,gpt4,t5}, 
recent works explore integrating LLMs inside the diffusion system to enhance prompt comprehension and representation~\cite{layoutgpt,lian2023llm,masterllm,sld,sda}. 
For instance, LMD~\cite{lian2023llm} performs layout-grounded image generation using captioned bounding boxes generated from prompts by LLMs~\cite{gpt4}. SLD~\cite{sld} introduces an LLM-driven image editing framework that utilizes open-set detectors~\cite{ovvit} and LLMs~\cite{gpt4} to identify misalignments and edit initial generation results. Additionally, recent studies explore the use of MLLMs to aid in image editing~\cite{MIGE,smartedit,pairdiffusion,specdiff,i2vedit} and customization~\cite{unimog,kosmosg,moma,easyref,finestyle,classdiffusion}. However, these methods primarily focus on executing human-provided instructions, rather than automatically correcting misalignment.

\vspace{1mm}
\noindent\textbf{Diffusion adapters}~\cite{dalle2,controlnet,t2iadapter,composer,unicontrol,cove,ipl,pfd,smoothdiffusion,ip} aim to extend the capability of diffusion models by incorporating additional input conditions beyond text prompts. Pioneering works like ControlNet~\cite{controlnet} and T2I-Adapter~\cite{t2iadapter} introduce structure-conditioned adapters to enable controllable image synthesis. On top of these baselines, works such as Composer and Uni-ControlNet~\cite{composer,unicontrol,unicontrol2} propose a unified adapter capable of handling various structural conditions. 
Recently, diffusion adapters enabling visual-encoding~\cite{ip,pfd,vcoder,brush2prompt,videobooth,coda} have continuously gained attention in variation, editing and video tasks. For instance, Prompt-free Diffusion~\cite{pfd} introduces the SeeCoder as an image context encoder, replacing the original text encoder to support image prompts. IP-Adapter~\cite{ip} integrates CLIP image encoder~\cite{clip} with decoupled cross-attention layers to create an effective image-prompt adapter. 

%% file: sec/3_1_method.tex
\vspace{-2mm}
\section{Methodology}\label{sec:3}
\label{sec:method}
\vspace{-1mm}
In this section, we first go through preliminaries of diffusion models with text and image prompts~\cite{sd,ip} in~\cref{sec:pre}. Then, we provide an in-depth explanation of our method: Implicit Multimodal Guidance (IMG) in~\cref{sec:edit,sec:ob,sec:correct}.

\subsection{Preliminaries}\label{sec:pre}
\textbf{Diffusion models}~\cite{sd,flux} are a class of generative methods involving forward and reverse processes. The forward process {is usually known as a gradual procedure, transforming the data point $\bm{x}_0$ into Gaussian noise $\bm{x}_T$ with ${T}$ steps.} {For example, a canonical formulation for $\bm{x}_t$ is defined as such:}
\begin{equation}
\setlength{\abovedisplayskip}{7pt}
\setlength{\belowdisplayskip}{7pt}
\begin{aligned}
    \bm{x}_t &= \alpha_t\bm{x}_{0} + \sigma_t\bm{\epsilon},
\end{aligned}
\label{eq:xt}
\end{equation}
where $t\sim\mathcal{U}(0,T)$, $\bm{\epsilon}\sim N(\bm{0}, \bm{I})$ is random Gaussian noise, $\alpha_t$ and $\sigma_t$ are predefined functions of $t$.
The reverse process iteratively transforms $\bm{x}_T$ into $\bm{x}_0$, assessing intermediate $\bm{x}_t$ through a well-trained deep neural net $\bm{\epsilon}_\theta$~\cite{ddpm,ddim,dpmslover,rf}.

Considering the text-to-image generation task, given an image $\bm{x}_0$ and text condition $\bm{c}_T$, the training objective of $\bm{\epsilon}_\theta$ is formulated as:
\begin{equation}
\setlength{\abovedisplayskip}{7pt}
\setlength{\belowdisplayskip}{7pt}
   L_{\text{diff}} = \mathbb{E}_{\bm{x}_0, \bm{\epsilon}, \bm{c}_T, t} \left\| \bm{\epsilon} - \bm{\epsilon}_\theta(\bm{x}_t, \bm{c}_T, t) \right\|_2^2.
\end{equation}

\begin{figure}[t]
    \centering
    \includegraphics[width=0.9\linewidth]{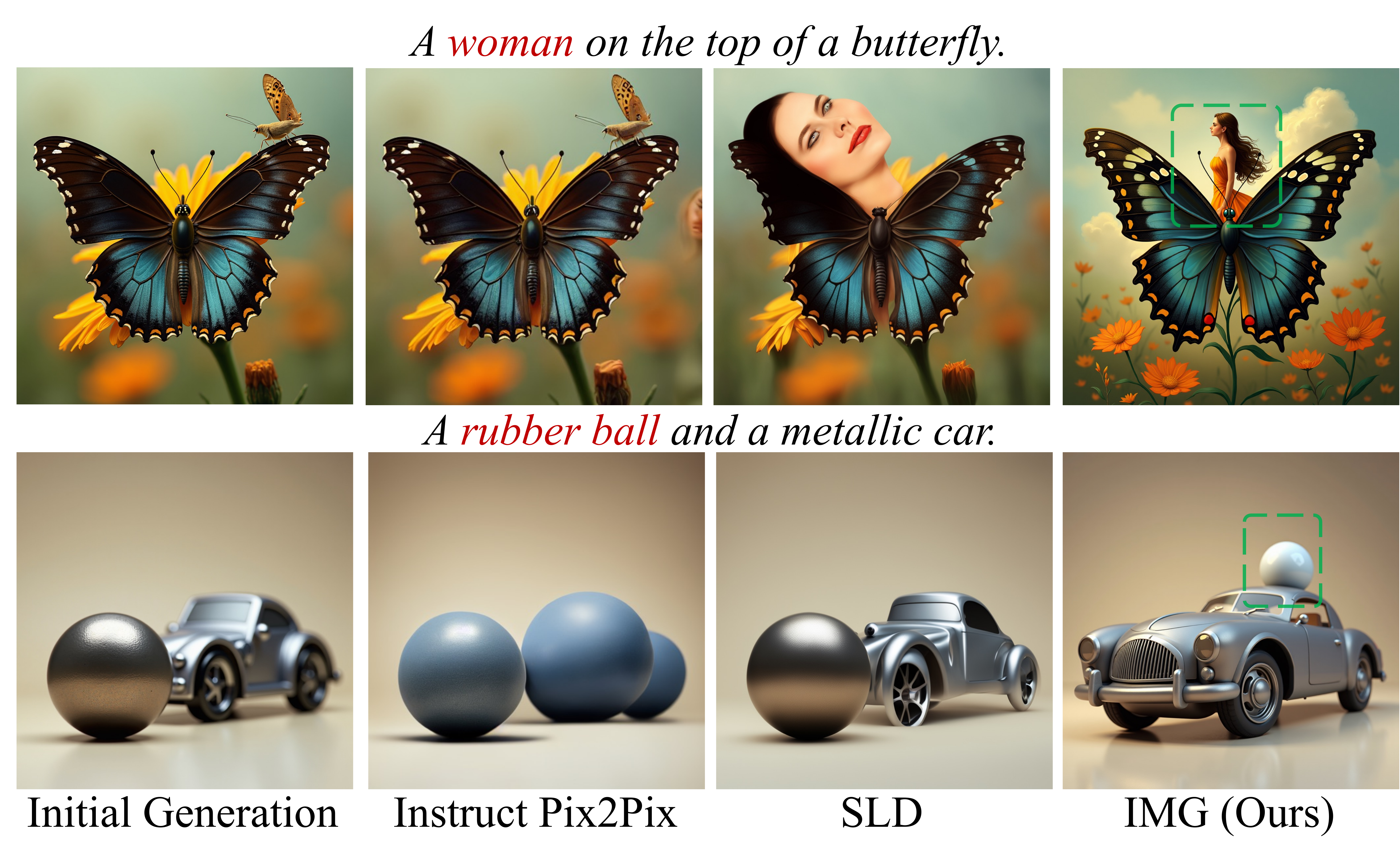}
      \vspace{-4mm}   
    \caption{\textbf{Comparison with editing-based methods.} We evaluate the performance of Instruct Pix2Pix~\cite{instructpix2pix} and SLD~\cite{sld} with IMG. For Instruct Pix2Pix, the instructions are ``add a woman'' and ``make the ball a rubber ball'', generated by our finetuned MLLM. }
    \label{fig:edit1}
    \vspace{-4mm}
\end{figure}

\noindent\textbf{Diffusion with image prompt}~\cite{ip,pfd,controlnet,unicontrol} has recently gained popularity in the community, in which diffusion models reconstruct images similar to the input image, usually through a frozen T2I model attaching an additional image prompt (IP) encoder that takes visual inputs. Thus, the training objective is extended to include an additional image condition $\bm{c}_I$, which is the encoded features of $\bm{x}_0$:
\begin{equation}
\setlength{\abovedisplayskip}{7pt}
\setlength{\belowdisplayskip}{7pt}
   L_{\text{diff}} = \mathbb{E}_{\bm{x}_0, \bm{\epsilon}, \bm{c}_T, \bm{c}_I, t} \left\| \bm{\epsilon} - \bm{\epsilon}_\theta(\bm{x}_t, \bm{c}_T, \bm{c}_I, t) \right\|^2_2. 
\end{equation}

During inference, users input both $\bm{c}_I$ and, optionally, $\bm{c}_T$ to create or enhance a content-consistent variant of $\bm{x}_0$. IP encoders~\cite{ip,xflux} are widely available for diffusion models such as SD series~\cite{sd} and FLUX~\cite{flux}.

%% file: sec/3_2_method.tex
\begin{figure*}[t]
    \centering
    \includegraphics[width=0.8\linewidth]{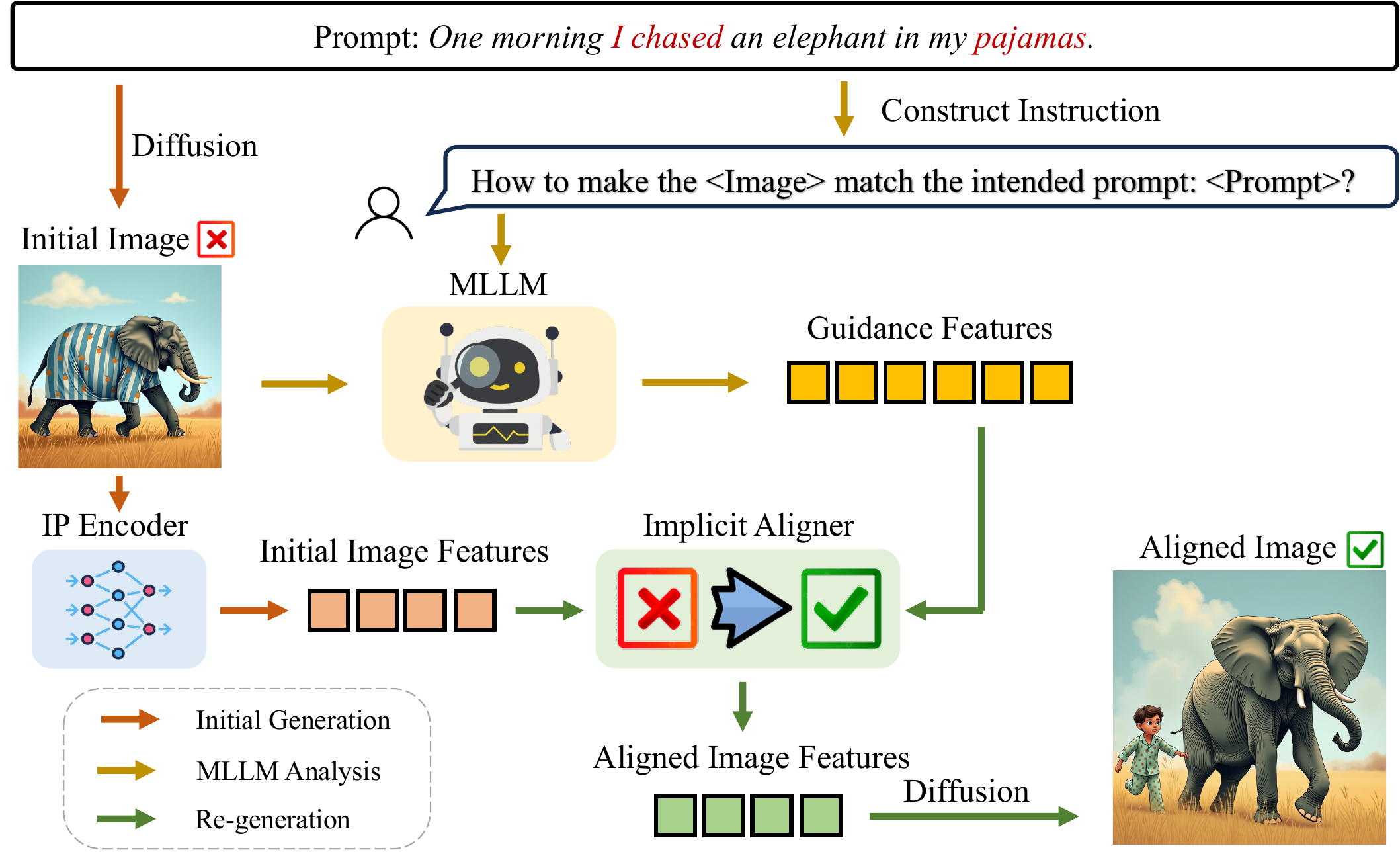}
      \vspace{-0mm}   
    \caption{\textbf{Overview of the Implicit Multimodal Guidance (IMG) framework.} Given an initial image that exhibits misalignments with its prompt, IMG begins by conducting an MLLM-driven misalignment analysis. Following this, IMG utilizes an Implicit Aligner to translate the initial image features into better-aligned features according to the MLLM's guidance. Finally, these aligned image features are incorporated as new conditions to re-generate images with improved prompt-image alignment.}
    \label{fig:architecture}
    \vspace{-3mm}
\end{figure*}

\subsection{MLLM-driven Misalignment Analysis}\label{sec:edit} 

{D}iffusion models sometimes misinterpret or overlook parts of a prompt when generating images. {To address this issue, our first step is to identify these misalignments via Multimodel Large Language Models (MLLMs) that are capable of visual question answering~\cite{vqa}.}

While MLLMs~\cite{llava} can describe image contents, there has been limited effort to customize them as misalignment detectors. In IMG, we address this by introducing a customized MLLM, which we finetune using instruction-based image data~\cite{instructpix2pix}, enabling the MLLM to analyze and respond to potential misalignments.

Specifically, the instruction-based dataset contains original images $I_0$, edited images $I_1$, prompts for both images $T_0$ \& $T_1$, and edit instructions $T_E$.
We utilize the data by taking out $I_0$, $T_1$ and $T_E$ as our training triplets.
While $I_0$ and $T_1$ are fed into the MLLM as inputs, we request the model to describe alignment through question prompts such as “How to make the $<$Original Image$>$ match the intended prompt: $<$Edited Prompt$>$?”, then supervise model outputs against $T_E$. 
Through experiments in~\cref{sec:abl}, we demonstrate that our specialized finetuning makes MLLM a much more reliable model for misalignment detection.
For more details regarding MLLM finetuning, please see our supplementary.

%% file: sec/3_3_method.tex
\subsection{{Implicit Multimodal Guidance}}\label{sec:correct}

Our next step in IMG is to improve the generated images by removing the previously detected misalignment. A straightforward baseline is to edit these images through existing editing methods like~\cite{instructpix2pix,sld}.
In~\cref{fig:edit1}, we provide a quick demo using Instruct Pix2Pix~\cite{instructpix2pix} and SLD~\cite{sld}, both showing unsatisfactory results.
For Instruct Pix2Pix, we use the responses generated by our finetuned MLLM as instructions. However, the result shows that the editing takes place incorrectly.
For SLD, although it successfully generates a woman in the first case, the overall aesthetic quality is degraded, and SLD fails to deal with the second case.

The unsuccessful try-out on existing editing methods motivates us to tackle the challenge from a new angle: set up an image re-generation process conditioned on \textit{implicit features}, highlighted as \textit{Implicit Multimodal Guidance} (IMG). A complete diagram of our IMG framework is shown in~\cref{fig:architecture}, in which a) we generate an initial image that may exhibit misalignments via a diffusion model; b) we detect potential misalignments via MLLM using questions same as those in the finetuning process, (\eg, ``How to make the $<$Image$>$ match the intended prompt: $<$Prompt$>$?''); and c) we re-generate the image conditioned on aligned features produced by the newly introduced Implicit Aligner.

Generally, our Implicit Aligner functions as an adapter network for the original diffusion model, facilitating the re-alignment of the initial image to better adhere to its prompt. Instead of explicitly editing the initial image~\cite{instructpix2pix,sld}, our design implicitly refines the diffusion conditioning features. Structured as a stack of cross-attention layers, the Implicit Aligner takes MLLM hidden states, namely guidance features, as one input and initial image features, embedded via an image prompt (IP) encoder, as the other. The guidance features promote modifications on initial features, producing better-aligned features that are trained to fit the embedded features of a well-aligned image. According to the properties of image prompt diffusion described in~\cref{sec:pre}, the diffusion model can then re-generate a more faithfully aligned image from these re-aligned features.

In the next section, we will introduce Iteratively Updated Preference Objective, the training objective of our Implicit Aligner that leverages human preference datasets~\cite{pick}.

%% file: sec/3_4_method.tex
\subsection{Iteratively Updated Preference Objective}\label{sec:ob}

As aforementioned, we leverage the same open-source human preference datasets~\cite{pick} adopted in finetuning-based alignment methods~\cite{dpo} to train our Implicit Aligner. 
These datasets consist of triplets $\{\bm{c}_T, \bm{x}_0^{w}, \bm{x}_0^{l}\}$, where $\bm{x}_0^{w}$ and $\bm{x}_0^{l}$ represent the human-preferred (winning) and non-preferred (losing) images based on the prompt $\bm{c}_T$ (\ie, $\bm{x}_0^{w} \succ \bm{x}_0^{l}$).
Let \(\bm{c}_I^l\) and \(\bm{c}_I^w\) denote the image prompt features of $\bm{x}_0^{l}$ and $\bm{x}_0^{w}$. 
The direction from \(\bm{c}_I^l\) to \(\bm{c}_I^w\) indicates a meaningful alignment-improving trajectory, conditioned on \(\bm{c}_T\)~\cite{ip,instantstyle}. 
Such trajectory suggests a basic objective for our Implicit Aligner, in which the network $f_\theta$ fits \(\bm{c}_I^w\) based on input conditions \(\bm{c} = (\bm{h}, \bm{c}_{I}^{l})\), where \(\bm{h}\) represents the MLLM guidance features and \(\bm{c}_I^l\) represents the image features, formulated as the following:
\begin{equation}
\setlength{\abovedisplayskip}{7pt}
\setlength{\belowdisplayskip}{7pt}
    \begin{aligned}
        &L_{\text{base}} = \mathbb{E}_{\bm{c}, \bm{c}_{I}^{w}} ||\bm{c}_{I}^{w} - f_{\theta}(\bm{c})||_2^2.
    \end{aligned}
    \label{eq:base}
\end{equation}

Besides the basics, we also draw inspiration from direct preference optimization (DPO)~\cite{dpollm} and self-play fine-tuning (SPIN)~\cite{spinllm} for enhanced feature alignment. In our case, we use DPO to finetune $f_{\theta}$ from a reference aligner $f_{\text{ref}}$ (a copy of $f_{\theta}$ from an earlier training iteration), encouraging the network to output $f_{\theta}(\bm{c})$ close to the preferred features \(\bm{c}_I^w\) while keeping distance from the non-preferred features \(\bm{c}_I^l\). SPIN also forces $f_{\theta}(\bm{c})$ to be close to \(\bm{c}_I^w\) while additionally forces $f_{\theta}(\bm{c})$ to be more preferred than $\bm{c}_{I}^{\text{ref}}=f_{\text{ref}}(\bm{c})$.
Adapting both DPO~\cite{dpollm,dpo} and SPIN~\cite{spinllm,spin}, we derive our objective as the following:
\begin{equation}
\setlength{\abovedisplayskip}{7pt}
\setlength{\belowdisplayskip}{7pt}
    \begin{aligned}
L_{\text{pref}} &= \mathbb{E}_{\bm{c}, \bm{c}_{I}^{w}, \bm{c}_{I}^{l}} \left[ \ell \left( \underbrace{\log \frac{p_\theta(\bm{c}_{I}^{w} \vert \mathbf{c})}{p_{\text{ref}}(\bm{c}_{I}^{w} \vert \mathbf{c})} - \log \frac{p_\theta(\bm{c}_{I}^{l} \vert \mathbf{c})}{p_{\text{ref}}(\bm{c}_{I}^{l} \vert \mathbf{c})}}_{\text{DPO}} \right.\right.\\
&\left.\left. + \underbrace{\log \frac{p_\theta(\bm{c}_{I}^{w} \vert \mathbf{c})}{p_{\text{ref}}(\bm{c}_{I}^{w} \vert \mathbf{c})} - \log \frac{p_\theta(\bm{c}_{I}^{\text{ref}} \vert \mathbf{c})}{p_{\text{ref}}(\bm{c}_{I}^{\text{ref}} \vert \mathbf{c})}}_{\text{SPIN}} \right) \right],\\
    \end{aligned}
    \label{eq:pref1}
\end{equation}
where $\ell$ is a monotonically decreasing convex function, usually implemented as $\ell(x)\colon=\log(1+e^{-x})$~\cite{dpo,spin}. Following~\cite{vae,learningreg,balancedmse}, we define $p_\theta$ and $p_{\text{ref}}$ as Gaussian distributions with means predicted by $f_{\theta}$ and $f_{\text{ref}}$ and a constant variance $\sigma$, \ie, $p(\bm{c}_{I}\vert \mathbf{c})=\mathcal{N}(\bm{c}_{I}; f(\mathbf{c}), \sigma)$. With some transformations, the objective in Eq.~\ref{eq:pref1} can be simplified to:
\begin{equation}
\setlength{\abovedisplayskip}{7pt}
\setlength{\belowdisplayskip}{7pt}
    \begin{aligned}
L_{\text{pref}} &= \mathbb{E}_{\bm{c}, \bm{c}_{I}^{w}, \bm{c}_{I}^{l}} [\ell(-[2(||\bm{c}_{I}^{w} - f_{\theta}(\bm{c})||_2^2 - ||\bm{c}_{I}^{w} - f_{\text{ref}}(\bm{c})||_2^2) \\
&- (||\bm{c}_{I}^{l} - f_{\theta}(\bm{c})||_2^2 - ||\bm{c}_{I}^{l} - f_{\text{ref}}(\bm{c})||_2^2)\\ &- ||f_{\text{ref}}(\bm{c}) - f_{\theta}(\bm{c})||_2^2])].
    \end{aligned}
    \label{eq:pref2}
\end{equation}

The detailed derivation of $L_{\text{pref}}$ can be found in the supplementary. The final training objective is then a combination of $L_{\text{base}}$ and $L_{\text{pref}}$ with a ratio parameter $\lambda$:
\begin{equation}
\setlength{\abovedisplayskip}{7pt}
\setlength{\belowdisplayskip}{7pt}
    \begin{aligned}
    L = L_{\text{base}} + \lambda L_{\text{pref}}.
    \end{aligned}
    \label{eq:sum}
\end{equation}

Unlike prior works such as DPO~\cite{dpo} and SPIN~\cite{spin}, which require a fixed reference net $f_{\text{ref}}$ with frozen weights, our Implicit Aligner can be trained from scratch and the reference net can be runtime updated. 
Specifically, we first randomly initialize $f_{\text{ref}}$ and later iteratively copy $f_{\theta}$ to $f_{\text{ref}}$ whenever $f_{\theta}$ outperforms $f_{\text{ref}}$.
In practice, we execute the substitution when $f_{\theta}(\bm{c})$ is closer to $\bm{c}_{I}^{w}$ than $f_{\text{ref}}(\bm{c})$ for $k$ consecutive iterations. Thereafter, we name our objective function $L$ the Iteratively Updated Preference Objective.

%% file: sec/4_exp.tex
\section{Experiments}\label{sec:exp}

\subsection{Experimental Setup}
\label{sec:setup}

\textbf{Baselines and models.}
Our experiments are conducted on two diffusion models: SDXL~\cite{sdxl} and FLUX.1 [dev] (FLUX)~\cite{flux}. We also compare IMG against Diffusion-DPO (SDXL-DPO)~\cite{dpo}, the top-performing finetuning-based alignment method, and SLD~\cite{sld}, the leading editing-based alignment method. We further compare IMG with leading compositional generation methods, ELLA~\cite{ella} and CoMat~\cite{Jiang2024CoMatAT}. For MLLM, we finetune LLaVA 1.5-13b~\cite{llava} on the Instruct-Pix2Pix dataset~\cite{instructpix2pix} for 1 epoch following~\cite{llava} and extract the last hidden layer features for guidance. We use the IP-Adapter~\cite{ip,xflux}, trained on SDXL and FLUX to enable image prompts and extract image features.

\begin{table*}[t]
    \centering
    \small
   {
    \resizebox{1\linewidth}{!}{\begin{tabular}{lcccccc|c|c}
        \toprule
        \multirow{2}{*}{Model} & \multicolumn{6}{c|}{Human Preference Datasets (HPD)} & \multirow{2}{*}{Parti-Prompts} & \multirow{2}{*}{User Study} \\
        & Anime & Concept-Art & Drawbench & Painting & Photo & Average & \\
        \midrule 
        SDXL~\cite{sdxl} &28.59&27.69&28.04&27.74&28.10& 28.03 & 27.71& \multirow{2}{*}{77.6\%} \\
        \rowcolor{gray!20}SDXL + IMG (Ours) &\textbf{29.14 (85.5\%)}&\textbf{28.31 (92.8\%)}&\textbf{28.48 (81.5\%)}&\textbf{28.33 (91.5\%)}&\textbf{28.49 (80.3\%)}&  \textbf{28.56 (87.2\%)} & \textbf{28.13 (79.2\%)}\\
        \midrule
        SDXL-DPO~\cite{dpo} &28.98&28.12&28.40&28.18&28.29& 28.39 & 27.99& \multirow{2}{*}{75.5\%}\\
        \rowcolor{gray!20}SDXL-DPO + IMG (Ours) &\textbf{29.42 (84.8\%)}&\textbf{28.55 (86.1\%)}&\textbf{28.68 (77.0\%)}&\textbf{28.58 (83.8\%)}&\textbf{28.66 (77.1\%)}& \textbf{28.79 (82.6\%)} & \textbf{28.34 (76.0\%)} \\
        \midrule
        FLUX~\cite{flux} &29.77 & 28.98 & 29.44  & 28.90  & 29.14  & 29.21 & 29.23& \multirow{2}{*}{66.9\%} \\
        \rowcolor{gray!20}FLUX + IMG (Ours)& \textbf{30.06 (67.0\%)} & \textbf{29.17 (62.0\%)}  & \textbf{29.80 (64.0\%)} &  \textbf{29.20 (69.5\%)}  & \textbf{29.41 (64.8\%)} & \textbf{29.48 (65.7\%)} & \textbf{29.61 (71.7\%)} \\
        \bottomrule
    \end{tabular}
    }}\vspace{-2mm}
    \caption{\textbf{Quantitative comparison with base models and finetuning-based alignment methods on HPD and Parti-Prompts}. We report the average HPS scores and the win rates of IMG over competing methods. Higher HPS scores indicate better preference alignment. Additionally, we conduct user studies to assess the real human preference rates of IMG.}\label{tab:hps}\vspace{-4mm}
\end{table*}

\vspace{1mm}
\noindent\textbf{Implementation details.} IMG's Implicit Aligner is trained on the Pick-a-Pic training set~\cite{pick} (the same as SDXL-DPO~\cite{dpo}) for 100K iterations with 8 A100 GPUs and a batch size of 8. The training data contains 851K preferred and non-preferred image pairs under specific prompts. We use the AdamW~\cite{adamw} optimizer with a constant learning rate of 1$\times10^{-4}$ and a weight decay of 1$\times10^{-4}$. The ratio parameter $\lambda$ in Eq.~\ref{eq:sum} is set to 1. The reference model updating step $k$ in \cref{sec:ob} is set to 10. We determine the optimal hyperparameters via the average Pick Score~\cite{pick} across generated images on 500 Pick-a-Pic test set prompts. Pick Score is a caption-aware alignment scoring model trained on Pick-a-Pic. For evaluation, we report the Human Preference Scores (HPS) on the Human Preference Datasets (HPD) test set~\cite{hps}, which includes 3,400 prompts across 5 categories and on the Parpi-Prompts~\cite{parti}, a diverse prompt dataset of 1,632 prompts ranging from brief concepts to complex sentences. We also report results on the T2I-CompBench~\cite{t2icomp}, which contains 1800 test prompts to validate compositional image generation capabilities.
We apply 50 sampling steps for SDXL and SDXL-DPO, and 30 for FLUX. More implementation details are provided in the supplementary.

\begin{figure}[t]
    \centering
    \includegraphics[width=\linewidth]{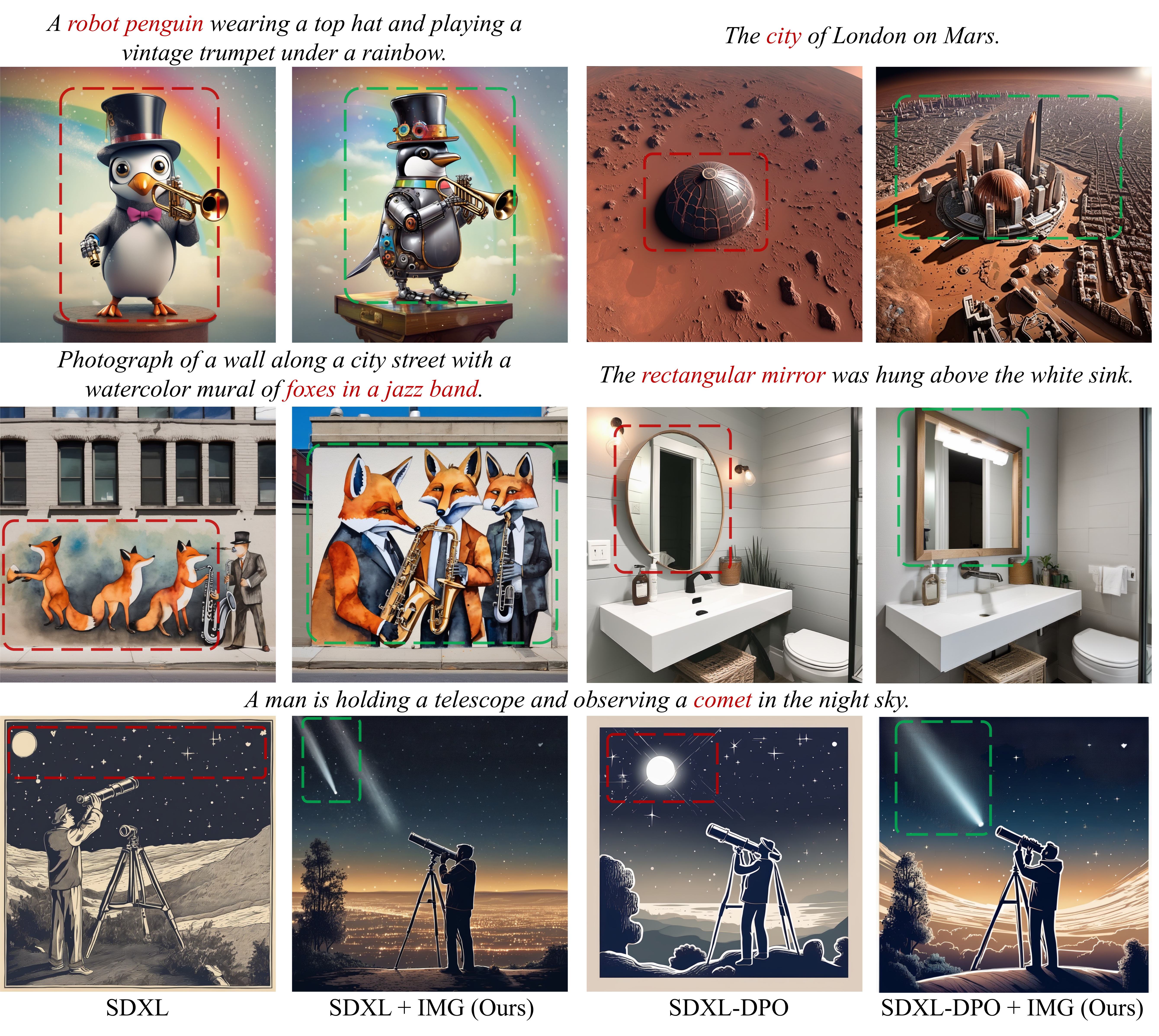}
      \vspace{-8mm}   
    \caption{\textbf{Qualitative comparison with base models and finetuning-based alignment methods.} The first two rows show that IMG addresses various misalignment types across different prompts, while the last row shows that IMG resolves misalignment issues that challenge both models.}
    \label{fig:main}\vspace{-4mm}
\end{figure}

\subsection{Comparison with Base Models and Finetuning-based Methods}

\noindent\textbf{Qualitative comparison.} In addition to comparing with FLUX~\cite{flux} in~\cref{fig:teaser}, we perform further quantitative evaluations by integrating IMG with the widely used SDXL~\cite{sdxl} model and its finetuned variant, SDXL-DPO~\cite{dpo}. The Implicit Aligner of IMG is trained using SDXL's image prompt encoder~\cite{ip} and seamlessly shared with SDXL-DPO.
As shown in~\cref{fig:main}, the results highlight a) IMG’s effectiveness in addressing diverse alignment issues from various aspects, such as concept comprehension (\eg, a penguin with a robotic body and foxes playing musical instruments), aesthetic quality (\eg, a well-constructed city on Mars), object addition (\eg, a comet in the sky) and object correction (\eg, a rectangular mirror), and b) IMG's flexibility to operate with SDXL-DPO without additional training. The last row also showcases cases where both SDXL and SDXL-DPO struggle, whereas IMG demonstrates clear improvements in alignment across both models.

\begin{table}[t]
    \centering
    \setlength{\tabcolsep}{3pt}
   {
    \resizebox{\linewidth}{!}{\begin{tabular}{lccccccc}
        \toprule
        \textbf{Model} & \textbf{Color↑} & \textbf{Shape↑} & \textbf{Texture↑} & \textbf{Spatial↑} & \textbf{Non-Spatial↑} & \textbf{Complex↑} \\
        \midrule 
        SDXL~\cite{sdxl}  & 0.5552 & 0.4878 & 0.5213 & 0.1915 & 0.3125 & 0.3403 \\
        SDXL-ELLA~\cite{ella}  & 0.7260 & \underline{0.5634} & \textbf{0.6686} & 0.2214 & 0.3069 & -\\
        SDXL-CoMat~\cite{Jiang2024CoMatAT}  & \textbf{0.7827}  & {0.5329}  & 0.6468  & \underline{0.2428}  & \underline{0.3187}  &\underline{0.3680}  \\
        SDXL-DPO~\cite{dpo} & 0.6957 & 0.5430 & 0.6428 & 0.2415 & 0.3160 & 0.3561 \\
        \rowcolor{gray!20}SDXL-DPO + IMG (Ours) & \underline{0.7294} & \textbf{0.5811} & \underline{0.6548} & \textbf{0.2484} & \textbf{0.3199} & \textbf{0.3874} \\
        \midrule
        FLUX~\cite{flux} & 0.7852 & 0.5633 & 0.6942 & 0.2792 & 0.3129 & 0.3891 \\
        \rowcolor{gray!20}FLUX + IMG (Ours) & \textbf{0.8060} & \textbf{0.5773} & \textbf{0.7008} & \textbf{0.2826} & \textbf{0.3139} & \textbf{0.3983} \\
        \bottomrule
    \end{tabular}}\vspace{-3mm}
    }
    \caption{\textbf{Quantitative comparison with base models and finetuning-based alignment methods on T2I-CompBench.} The best results are in \textbf{bold} and the second-best results are \underline{underlined}.}~\label{tab:t2i}\vspace{-6mm}
\end{table}

\vspace{1mm}
\noindent\textbf{Quantitative comparison.}
In~\cref{tab:hps}, we conduct quantitative evaluations on the HPD ~\cite{hps} and Parti-Prompts~\cite{parti} datasets, which contain thousands of prompts across diverse categories and complexities. We evaluate images generated by the base models and those re-generated by IMG, using HPS scores~\cite{hps}, and report the win rates of IMG over the base models. The results demonstrate that IMG serves as a general framework that consistently enhances alignment across different base models. Notably, IMG shares the same training data as SDXL-DPO, yet when integrated with SDXL, it outperforms SDXL-DPO and achieves an average win rate of 84.6\% over SDXL. Furthermore, when integrated with SDXL-DPO, IMG achieves even higher performance, with an average win rate of 80.5\% over SDXL-DPO. This suggests that IMG not only enhances alignment more effectively but also complements finetuning-based methods boosting their performance without requiring additional data. When combined with the state-of-the-art FLUX model, IMG also achieves a strong alignment performance with an average win rate of 67.6\%. In addition, we conduct user studies where 33 evaluators were asked to do an A-B test on 30 random image pairs generated by each base model and IMG with the same prompt. Each unique pair was assessed by 3 evaluators, and only fully consistent votes were used to compute the final win rates. Results show that approximately 70\% of users prefer our re-aligned images over the originals. 
In~\cref{tab:t2i}, beyond general prompt sets, we further assess IMG on the compositional image generation benchmark, T2I-CompBench~\cite{t2iadapter}. Without specialized training on compositional prompts like CoMat~\cite{Jiang2024CoMatAT}, IMG operates in a zero-shot manner while achieving leading performance with both SDXL and FLUX.

\subsection{Comparison with Editing-based Methods.}

\begin{figure}[t]
    \centering
    \includegraphics[width=1.0\linewidth]{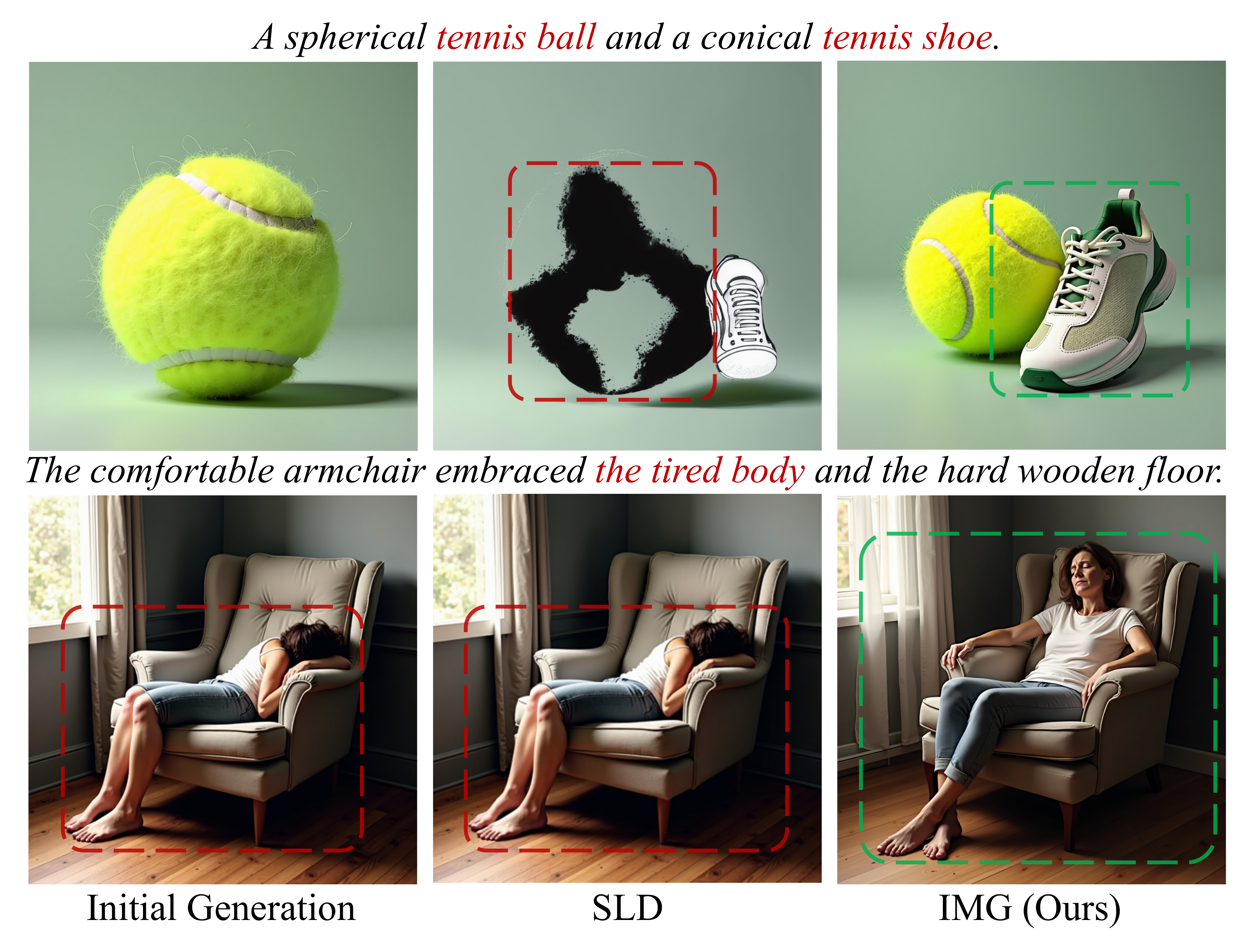}
      \vspace{-8mm}   
    \caption{\textbf{Qualitative comparison with editing-based methods.} IMG surpasses SLD in visual quality and image comprehension.}\vspace{-4mm}
    \label{fig:edit-main}
\end{figure}

\noindent\textbf{Qualitative comparison.}
In~\cref{fig:edit-main}, we compare IMG with the leading editing-based alignment method, SLD~\cite{sld}. In the first case, although SLD recognizes the missing tennis shoe, it incorrectly deletes the intended tennis ball. In the second case, SLD fails to recognize the unintended human gesture, resulting in no edits. These limitations arise primarily because the LLM in SLD interprets image content through text-based bounding box descriptions provided by a detector,  which introduces several risks: a) inaccurate detection results that lead to incorrect editing instructions from the LLM (\eg, removing the tennis ball) and b) an inability to address quality-related misalignments (\eg, overlooking the unintended gesture). In contrast, IMG, leveraging an MLLM that processes both images and prompts as input, more accurately identifies potential misalignments.

\begin{table}[t]
\centering
\small
\resizebox{\linewidth}{!}{\begin{tabular}{lccc}
\toprule
 & SDXL & SDXL-DPO  & FLUX  \\
\midrule
Base Model   & 28.03 & 28.39 & 29.21  \\
Base Model + SLD~\cite{sld}  & 27.35 & 27.52 & 28.14  \\
\rowcolor{gray!20}Base Model + IMG (Ours) &\textbf{28.56} & \textbf{28.79} &\textbf{29.48} \\
\bottomrule 
\end{tabular}}\vspace{-2mm}
\caption{\textbf{Quantitative comparison with editing-based methods.} SLD~\cite{sld} is designed to enhance alignment through local editing but often at the cost of overall image quality.}\label{tab:editscore}\vspace{-6mm}
\end{table}

\vspace{1mm}
\noindent\textbf{Quantitative Comparison.} In~\cref{tab:editscore}, we further compare IMG with SLD on the HPD benchmark. The results reveal that SLD leads to performance degradation, aligning with our qualitative observations and analysis. While SLD may enhance alignment in locally edited regions, the accumulated errors from the detector, LLM, and editing pipeline ultimately reduce overall image quality and alignment performance.
In contrast, IMG improves alignment through a native and more reliable re-generation process.

\subsection{Discussion}\label{sec:dis}

\vspace{1mm}
\noindent\textbf{MLLM misalignment detection accuracy.} 
We evaluate our finetuned MLLM as follows: a) We sample 300 instruction-based editing cases from SEED-Data-Edit~\cite{seed} as the test set; b) Both the original and the finetuned MLLM detect misalignment by predicting editing instructions; and c) GPT-4~\cite{gpt4} is employed to verify whether the predictions are semantically consistent with the ground truth. As a result, the finetuned MLLM achieves a 76.3\% ``yes'' rate, compared to 42.3\% for the original MLLM. 

\vspace{1mm}
\noindent\textbf{Generalization to different MLLMs.} In~\cref{tab:qwen}, we additionally integrate IMG with Qwen2.5-VL-7B. Evaluations on the HPD benchmark show consistent alignment improvements across different MLLMs, highlighting IMG's strong generalization capability.

\begin{figure}[t]
    \centering
    \includegraphics[width=0.965\linewidth]{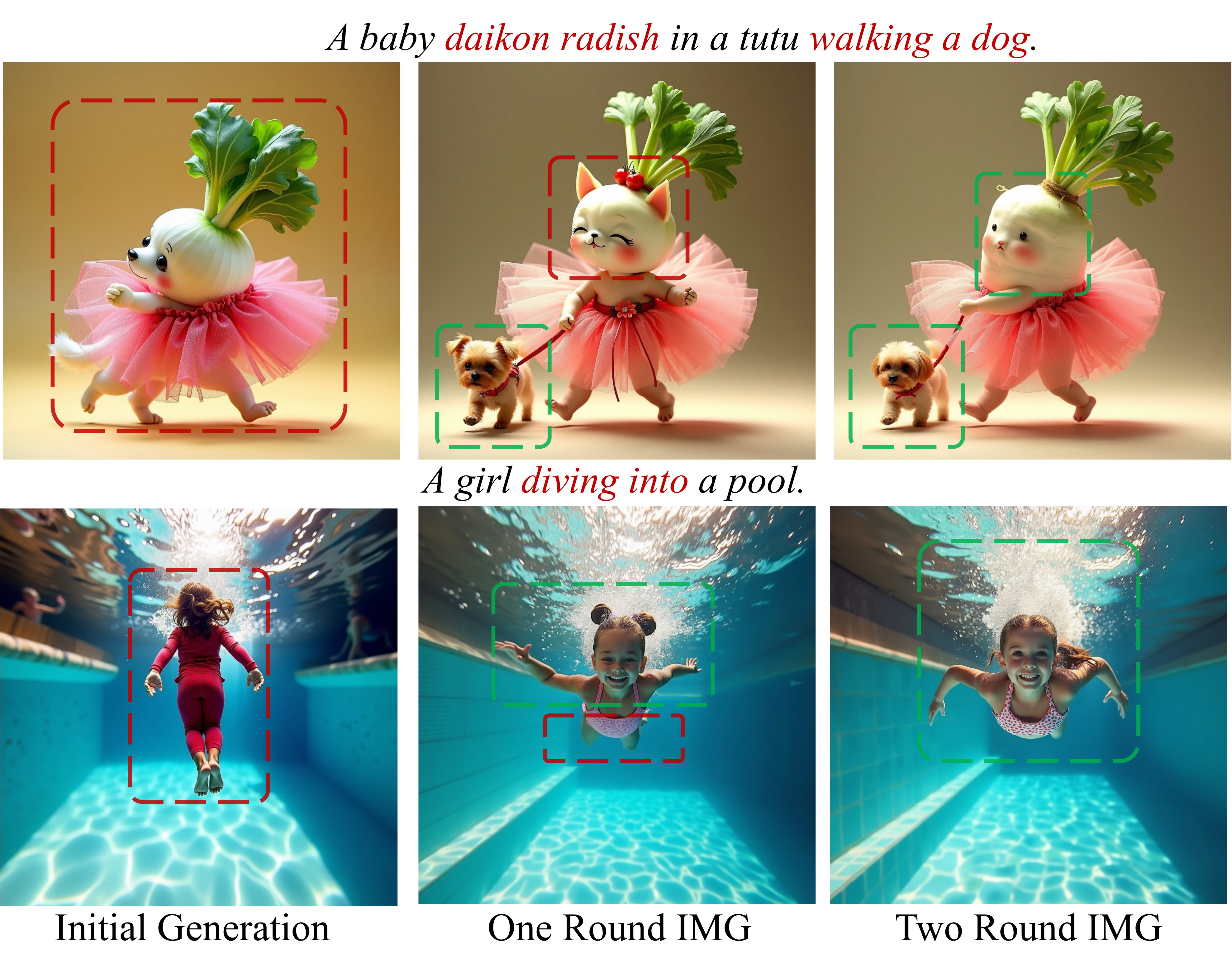}
      \vspace{-3.5mm}   
    \caption{\textbf{Multi-round generation results.} IMG continuously improves prompt-image alignment by executing multiple rounds.}
    \label{fig:multi}
\end{figure} 

\begin{table}[t]
\centering
\small
\resizebox{\linewidth}{!}{\begin{tabular}{lccc}
\toprule
 & SDXL & SDXL-DPO  & FLUX  \\
\midrule
Base Model   & 28.03 & 28.39 & 29.21  \\
\rowcolor{gray!20}Base Model + IMG (LLaVA) &\textbf{28.56} & \textbf{28.79} &\textbf{29.48} \\
\rowcolor{gray!20} Base Model + IMG (Qwen-VL) &\textbf{28.48} & \textbf{28.73} &\textbf{29.52} \\
\bottomrule 
\end{tabular}}\vspace{-2.5mm}
\caption{\textbf{Generalization capability to different MLLMs.} We further integrate IMG with Qwen-VL~\cite{qwenvl}. The results indicate consistent improvement across different MLLMs.}\label{tab:qwen}\vspace{-4mm}
\end{table}

\begin{figure*}[t]
    \centering
    \includegraphics[width=\linewidth]{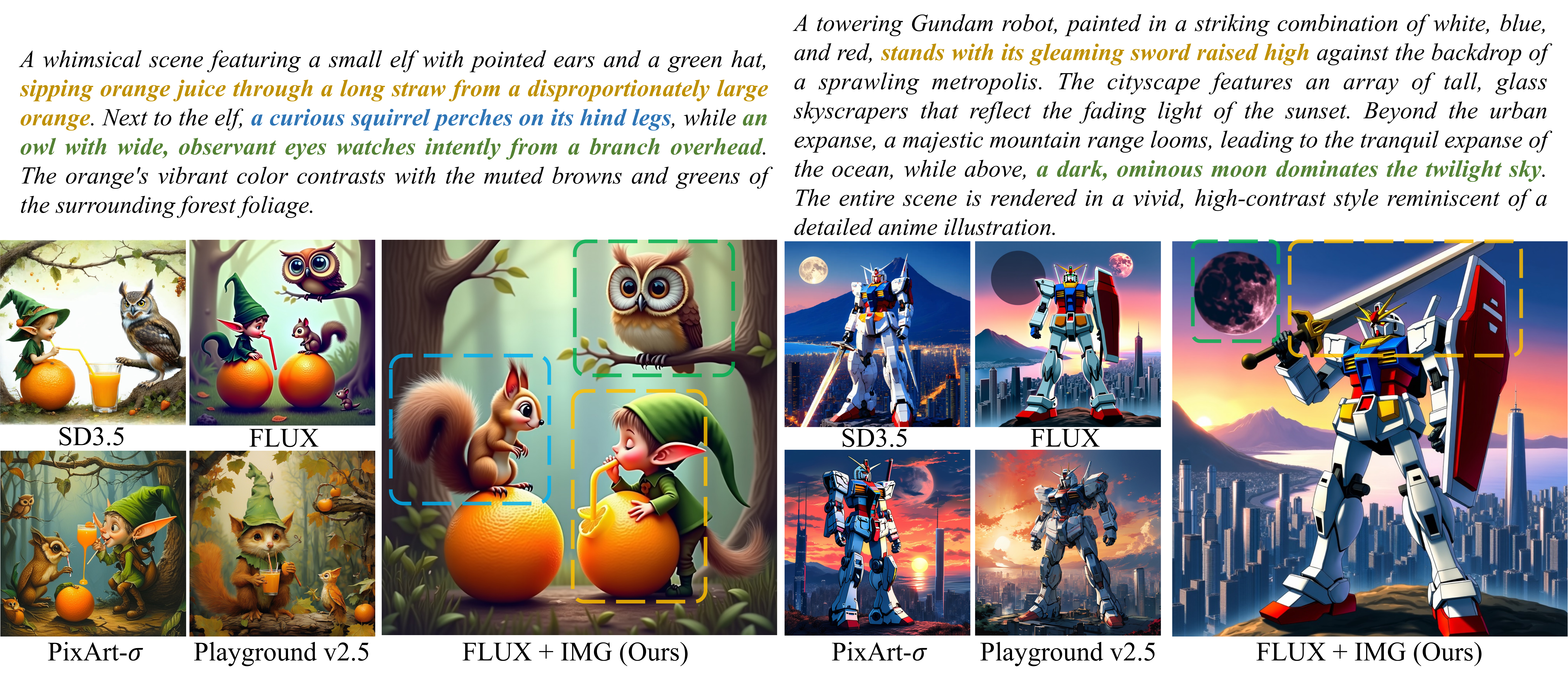}
      \vspace{-9mm}   
    \caption{\textbf{Dense prompt generation.} We integrate IMG with FLUX~\cite{flux} and compare it against leading community models, including Stable Diffusion 3.5 Large (SD3.5)~\cite{sd35}, PixArt-$\sigma$~\cite{pixartsigma} and Playground v2.5~\cite{pg25}, using complex dense prompts.}
    \label{fig:long}
\end{figure*}

\begin{table*}[t]
    \centering
    \begin{subtable}[t]{0.36\textwidth}
        \centering
        \resizebox{\linewidth}{!}{\begin{tabular}{ccc}
            \toprule
            Guidance features & SDXL & SDXL-DPO \\ \midrule 
            Base Model & 22.34 & 22.57\\
            \midrule
            No Guidance & 22.22 & 22.39  \\
             Text Embedding & 22.38 & 22.68 \\ 
             Original MLLM & 22.59 & 22.81 \\ 
             \rowcolor{gray!20}Finetuned MLLM (IMG) & \textbf{22.74} & \textbf{22.98} \\ 
             \bottomrule
        \end{tabular}}
        \caption{Impact of different guidance features.}
        \label{tab:subtable1}
    \end{subtable}
    \hfill
    \begin{subtable}[t]{0.269\textwidth}
        \centering
        \resizebox{\linewidth}{!}{\begin{tabular}{ccc}
            \toprule
            Ratio & SDXL & SDXL-DPO \\ \midrule
            Base Model & 22.34 & 22.57\\
            \midrule
            0 & 22.56 & 22.76\\ 
             0.5 & 22.56 & 22.80 \\ 
             \rowcolor{gray!20}1 (IMG) & \textbf{22.74} & \textbf{22.98} \\ 
             2 & 22.50 & 22.79 \\ 
             \bottomrule
        \end{tabular}}
        \caption{Impact of different $\lambda$ values in Eq.~\ref{eq:sum}.}
        \label{tab:subtable3}
    \end{subtable}
    \hfill
    \begin{subtable}[t]{0.32\textwidth}
        \centering
        \resizebox{\linewidth}{!}{\begin{tabular}{ccc}
            \toprule
            Components & SDXL & SDXL-DPO \\ \midrule
            Base Model & 22.34 & 22.57\\ 
            \midrule
             + $L_{\rm base}$ & 22.56 & 22.76 \\ 
             + $L_{\rm pref-DPO}$ & 22.65 & 22.88 \\ 
             \rowcolor{gray!20}+ $L_{\rm pref-SPIN}$ (IMG) & \textbf{22.74} & \textbf{22.98} \\ 
             \bottomrule
        \end{tabular}}
        \caption{Impact of each loss component.}
        \label{tab:subtable2}
    \end{subtable}
    \vspace{-3mm}
    \caption{\textbf{Ablation studies.} We examine the impact of different (a) guidance features, (b) loss ratios, and (c) loss components on the Pick-a-Pic test set via the average Pick Score. Higher Pick Scores indicate better preference alignment.}
    \label{ablation}\vspace{-3mm}
\end{table*}

\vspace{1mm}
\noindent\textbf{Multi-round generation.} In~\cref{fig:multi}, we demonstrate that IMG functions as an iterable framework that continuously enhances alignment through multiple rounds. For details that are not fully aligned in the first round IMG, \eg, the dog-like shape of the daikon radish and the action of diving, a second round of IMG further refines these details.

\vspace{1mm}
\noindent\textbf{Dense prompt generation.} Generating images from long, complex prompts remains a challenging task for diffusion models~\cite{parti,ella,pg25}. In~\cref{fig:long}, we integrate IMG with FLUX and compare it against leading community models, including Stable Diffusion 3.5 Large (SD3.5)~\cite{sd35}, PixArt-$\sigma$~\cite{pixartsigma} and Playground v2.5~\cite{pg25}. The results demonstrate that IMG substantially enhances details in generated images (as highlighted by the colored texts and boxes), enabling FLUX to stand out among competitors.

\subsection{Ablation Studies}\label{sec:abl}

In~\cref{ablation}, we examine the impact of different implicit guidance features, loss components, and loss ratios on the Pick-a-Pic test set via Pick Score to determine the optimal training scheme and hyperparameters. 

\vspace{1mm}
\noindent\textbf{Impact of different guidance features.} In~\cref{tab:subtable1}, we examine various guidance features used by the Implicit Aligner to refine initial image features into better-aligned features. ``No Guidance", which directly uses initial image features as aligned ones, performs worse than the ``Base Model", underscoring the need for feature alignment. Using ``Text Embedding" (text prompt embeddings from the diffusion text encoder) to guide feature alignment provides only minor improvements, as it lacks information on specific misalignments. ``Original MLLM" features offer significant gains across two base models, making SDXL + IMG competitive with SDXL-DPO and highlighting MLLM’s alignment capability. IMG achieves the best alignment scores with our ``Finetuned MLLM''. This aligns with the accuracy improvement in~\cref{sec:dis} and validates the effectiveness of our customized finetuning task in~\cref{sec:edit}.

\vspace{1mm}\noindent\textbf{Impact of different $\lambda$ values.} In~\cref{tab:subtable3}, we examine the impact of varying the loss ratio $\lambda$ in Eq.~\ref{eq:sum}, which controls the strength of $L_{\rm pref}$. Setting $\lambda=0$ reduces the objective to $L_{\rm base}$ alone. We find that a small $\lambda$ may not sufficiently activate the effect of $L_{\rm pref}$, while an excessively large $\lambda$ can lead to training instability and thus suboptimal performance, as the Implicit Aligner is trained from scratch rather than finetuned from a reference model. Empirically, $\lambda=1$ provides a balanced and effective trade-off.

\vspace{1mm}\noindent\textbf{Impact of each loss component.} In~\cref{tab:subtable2}, we examine each loss component in Eq.~\ref{eq:sum}. Using the basic objective $L_{\rm base}$ (Eq.~\ref{eq:base}), SDXL + IMG already achieves alignment performance competitive with SDXL-DPO. Furthermore, IMG is compatible with SDXL-DPO, enabling additional performance gains. For $L_{\rm pref}$, we separately evaluate the impact of its DPO and SPIN components in Eq.~\ref{eq:pref1}. Results demonstrate that these two components provide progressive performance improvements in alignment.

%% file: sec/5_conclusion.tex
\section{Conclusion}

In this paper, we propose Implicit Multimodal Guidance (IMG), a novel re-generation-based alignment framework. Unlike existing finetuning-based and editing-based approaches, IMG enhances alignment performance without requiring additional finetuning data or explicit editing operations. 
Specifically, given a generated image and its prompt, IMG involves an MLLM that identifies potential misalignments and an Implicit Aligner that reduces misalignments and facilitates re-generation by refining diffusion conditioning features. The Implicit Aligner is optimized through a trainable Iteratively Updated Preference Objective. Extensive qualitative and quantitative evaluations on SDXL, SDXL-DPO, and FLUX show that IMG outperforms existing alignment methods. Furthermore, IMG acts as a flexible plug-and-play adapter, seamlessly enhancing prior finetuning-based alignment methods. 

\section*{Acknowledgments}
\label{sec:ack}

This research was supported in part by National Science Foundation under Award \#2427478 - CAREER Program, and by National Science Foundation and the Institute of Education Sciences, U.S. Department of Education under Award \#2229873 - National AI Institute for Exceptional Education. This project was also partially supported by cyberinfrastructure resources and services provided by College of Computing at the Georgia Institute of Technology, Atlanta, Georgia, USA.

%% file: sec/appendix.tex
\clearpage
\setcounter{page}{1}
\maketitlesupplementary
\appendix

\section{Implementation Details}

\subsection{Baselines and Models.}
\begin{figure}[h]
  \centering
  \includegraphics[width=0.7\linewidth]{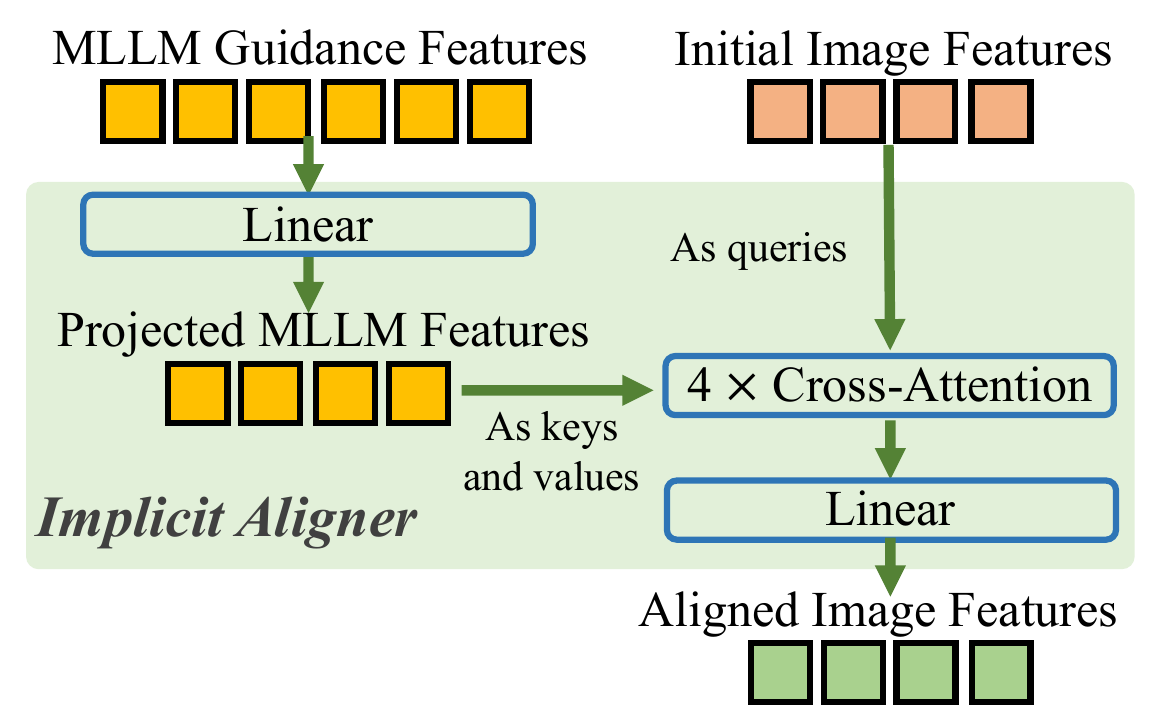}
  \caption{\textbf{Detailed architecture of Implicit Aligner.} Our Implicit Aligner contains 4 cross-attention layers and 2 linear layers. The number of color cubes here represents the token dimensions rather than the number of tokens.}
  \label{fig:supparchi}
\end{figure}

\begin{figure}[h]
    \centering
    \includegraphics[width=\linewidth]{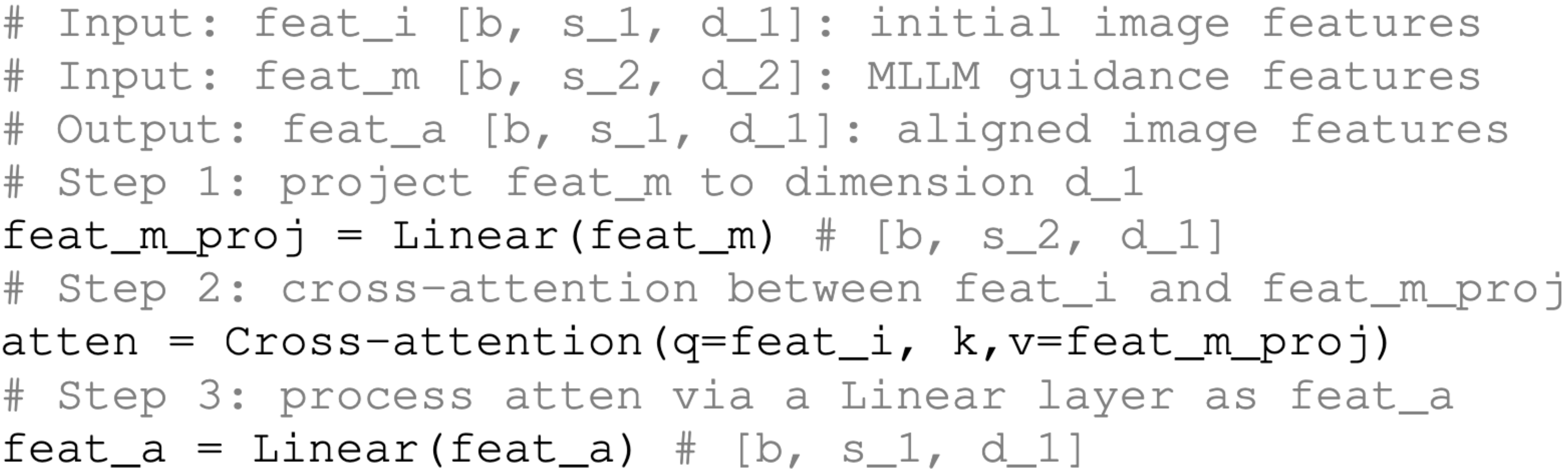}
    \caption{\textbf{Pseudo code of Implicit Aligner.} Our Implict Aligner (1) {projects MLLM features} to the same dimension as image features; (2) {conducts cross-attention} between initial image features and projected MLLM features; and (3) {processes attention outputs} with a linear layer as {aligned image features}.}
    \label{fig:suppcode}\vspace{-1mm}
\end{figure}

Our experiments are based on two diffusion models: SDXL~\cite{sdxl}, a widely adopted base diffusion model for alignment tasks, and FLUX.1 [dev] (FLUX)~\cite{flux}, a recent state-of-the-art flow-matching-based diffusion transformer. To compare with finetuning-based methods, we use the top-performing finetuned variant of SDXL, SDXL-DPO, which applies the Diffusion-DPO~\cite{dpo} technique, demonstrating the superiority of IMG and its compatibility with finetuning-based methods. For comparison with editing-based methods, we adopt the leading SLD as our baseline to highlight the advantages of IMG in visual comprehension and aesthetic quality. We further compare IMG with leading compositional generation methods, ELLA~\cite{ella} and CoMat~\cite{Jiang2024CoMatAT}, to evaluate the compositional generation capabilities. For MLLM, we finetune LLaVA 1.5-13b~\cite{llava} on the Instruct-Pix2Pix dataset~\cite{instructpix2pix} for 1 epoch, using the finetuning task format shown in \cref{fig:MLLMft}, and extract features from the last hidden layer for guidance. We use the IP-Adapter~\cite{ip,xflux}, trained on SDXL and FLUX, to enable image prompts and extract image features, with the image prompt scale set to 0.2. The Implicit Aligner takes both MLLM and image features as input and is implemented as a stack of 4 cross-attention layers and 2 linear layers. A detailed illustrative diagram of Implicit Aligner is shown in~\cref{fig:supparchi}, accompanied by its execution pseudo code in~\cref{fig:suppcode}. During inference, we execute the forward pass 3 times to enhance performance. 

\subsection{Datasets and Benchmarks.} 
For Implicit Aligner training, we use the same Pick-a-Pic training set~\cite{pick} as Diffusion-DPO~\cite{dpo}, which consists of 851K pairs of preferred and unpreferred images generated under specific prompts. The preference labels are annotated by human observers. To determine the optimal training scheme and hyperparameters, we conduct ablation studies by evaluating the average Pick Score~\cite{pick} across generated images using 500 unique prompts from the Pick-a-Pic test set. The Pick Score is a caption-aware preference scoring model trained on Pick-a-Pic. For evaluation, we report Human Preference Scores v2 (HPS v2) across generated images on the Human Preference Datasets v2 (HPD v2) test set~\cite{hps}, which includes 3,400 prompts across five categories, as well as the Parpi-Prompts~\cite{parti}, a diverse dataset of 1,632 prompts ranging from brief concepts to complex sentences. HPS v2 is a caption-aware preference scoring model trained on HPD v2. We also report results on the T2I-CompBench~\cite{t2icomp}, which contains 1800 test prompts to validate compositional image generation capabilities. For each test in user studies, 33 evaluators were asked to do an A-B test on 30 random image pairs generated by the base model and IMG with the same prompt. Each unique pair was assessed by 3 evaluators, and only fully consistent votes were used to compute the final win rates.
For MLLM finetuning, we extract triplets of $\{$Original Image, Edited Prompt, Edit Instruction$\}$ from the CLIP-filtered Instruct-Pix2Pix dataset~\cite{instructpix2pix}, which contains 313K samples.

\subsection{MLLM Finetuning.} 
\begin{figure}[t]
    \centering
    \includegraphics[width=\linewidth]{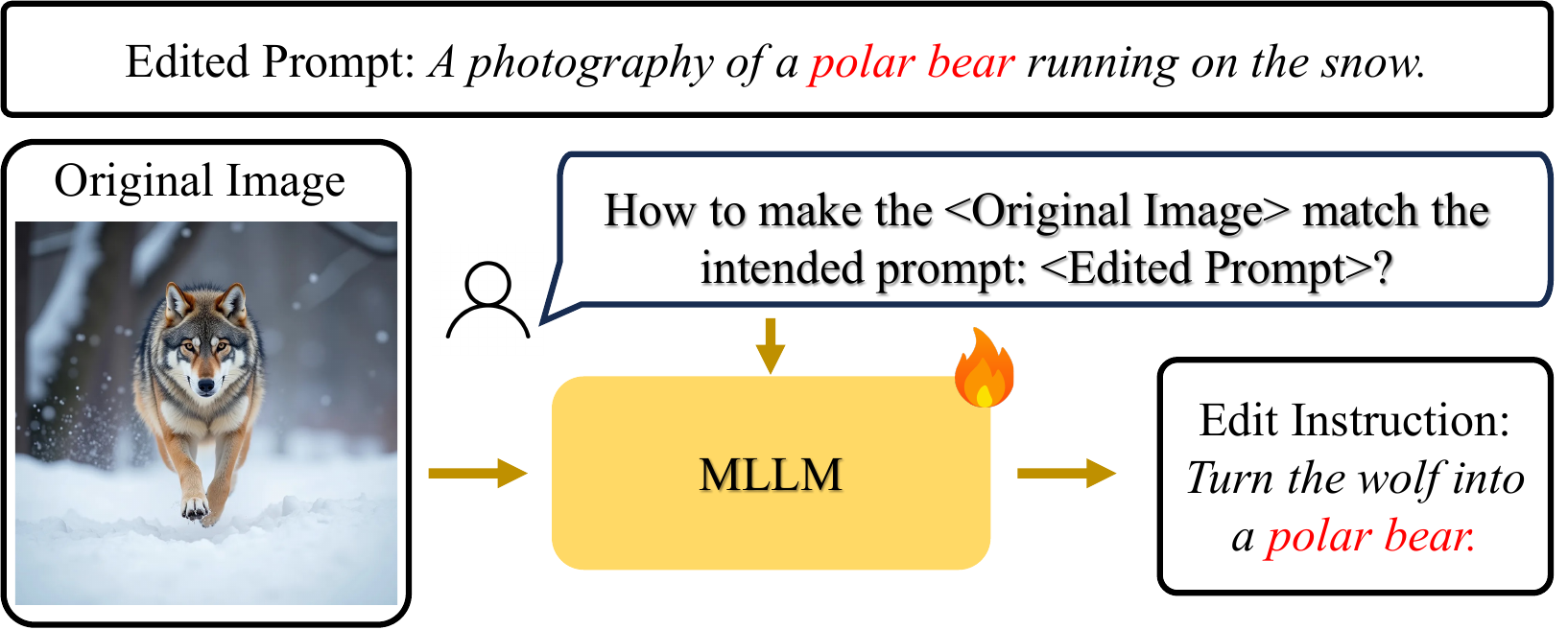}
      \vspace{-5mm}   
    \caption{\textbf{MLLM finetuning on instruction-based image data.} We conduct finetuning on $\{$Original Image, Edited Prompt, Edit Instruction$\}$ triplets from image editing datasets~\cite{instructpix2pix} to enhance MLLM's comprehension on prompt-image misalignments.}
    \label{fig:MLLMft}
    \vspace{-2mm}
\end{figure}

\begin{figure}[t]
    \centering
    \includegraphics[width=0.8\linewidth]{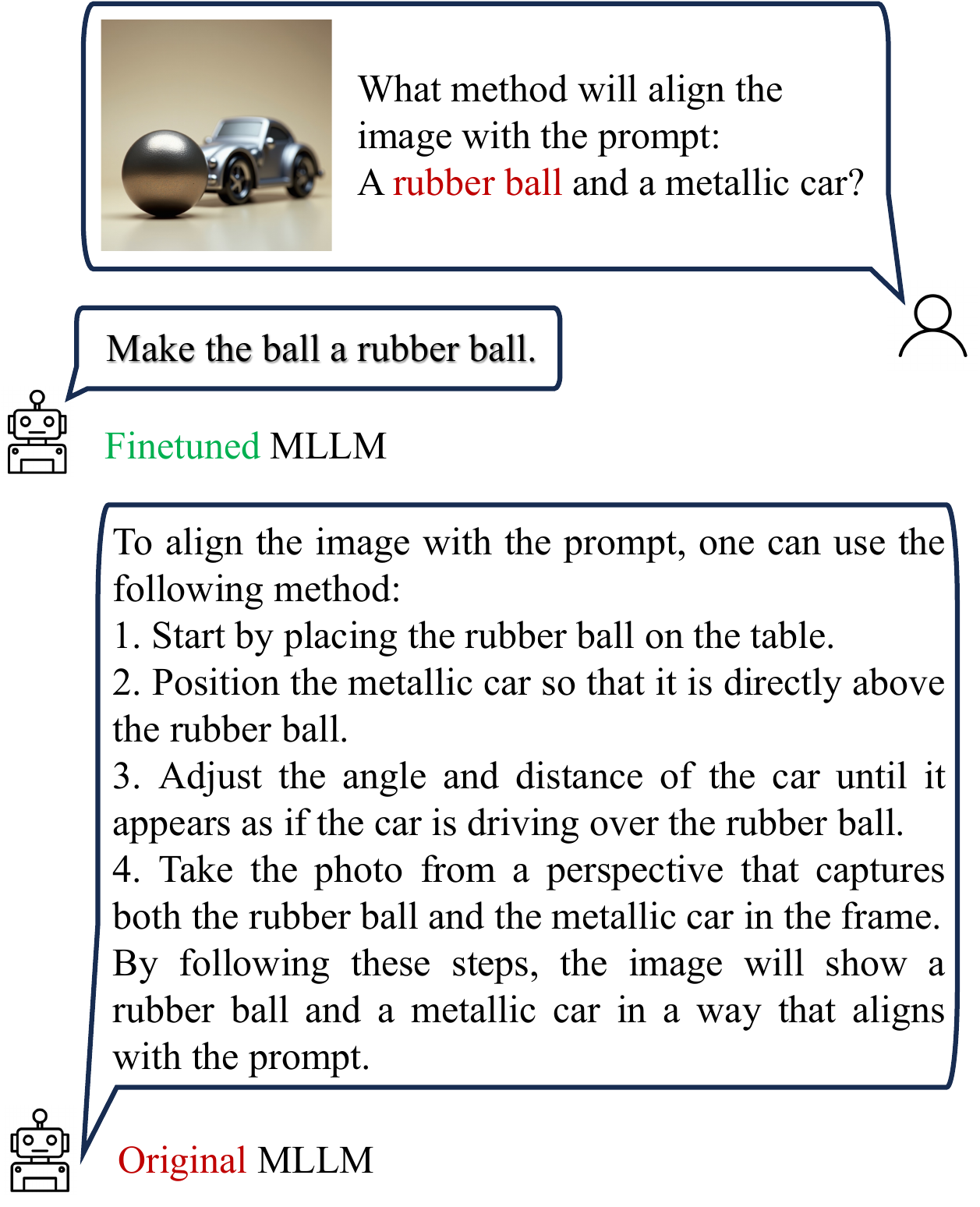}
      \vspace{-4mm}   
    \caption{Text response comparison of the original MLLM and our finetuned MLLM. The original MLLM primarily outlines an image generation process based on the prompt, while our finetuned MLLM emphasizes aligning the input image with the provided prompt, showcasing its misalignment detection capability.}\vspace{-5mm}
    \label{fig:mllm_supp}
\end{figure}

To customize a pretrained MLLM as a misalignment detector, we finetune LLaVA 1.5-13b~\cite{llava} on the Instruct-Pix2Pix dataset~\cite{instructpix2pix} for 1 epoch. We use training triplets consisting of original images $I_0$, edited prompts $T_1$, and edit instructions $T_E$. While $I_0$ and $T_1$ are fed into the MLLM as inputs, we prompt the model to describe the alignment by asking questions such as, 'How can the $<$Original Image$>$ match the intended prompt: $<$Edited Prompt$>$?', and supervise the model's outputs against $T_E$ (see~\cref{fig:MLLMft}). To prevent overfitting, we randomly select one of 100 different misalignment detection questions for each sample. The fine-tuning hyperparameters follow the standard configurations in~\cite{llava}. In~\cref{fig:mllm_supp}, we compare the text responses of the original MLLM and our fine-tuned MLLM. The original MLLM primarily outlines an image generation process based on the prompt, while our finetuned MLLM emphasizes aligning the input image with the provided prompt, showcasing its misalignment detection capability.

\subsection{IMG Training and Evaluation.} 

Our Implicit Aligner is trained on the Pick-a-Pic training set~\cite{pick} for 100K iterations with 8 A100 GPUs and a batch size of 8. We use the AdamW~\cite{adamw} optimizer with a constant learning rate of 1$\times10^{-4}$ and a weight decay of 1$\times10^{-4}$. The ratio parameter in Eq.~\ref{eq:sum} is set to 1. The reference model updating step $k$ in \cref{sec:ob} is set to 10. The training process takes about 10-15 hours. For evaluation, we set classifier-free guidance~\cite{cfg} to 7.5 for SDXL and SDXL-DPO, and 3.5 for FLUX. 
Sampling steps are set to 50 for SDXL and SDXL-DPO, and 30 for FLUX. The MLLM in IMG consumes about 4\% additional inference time and 15G (Qwen-VL-7B) - 25G (LLaVA-13B) GPU memory.

\section{Objective Derivation}

This section presents the detailed derivation of our proposed Iteratively Updated Preference Objective $L$ in~\cref{sec:ob}, which is a combination of a basic objective $L_{\text{base}}$ and a preference objective $L_{\text{pref}}$. To enhance generality and clarity, we substitute the $\bm{c}_I^w$ and $\bm{c}_I^l$ in~\cref{sec:ob} with more general forms, $\bm{x}_{w}$ and $\bm{x}_{l}$. These denote the preferred and non-preferred outputs of a regression model $f_{\theta}$ (the Implicit Aligner in IMG), under a given condition $\bm{c}$. In essence, the training procedure operates on triplets $\{\bm{c}, \bm{x}_{w}, \bm{x}_{l}\}$.

\subsection{Basic Objective}\label{sec:bo}
The primary goal of $f_{\theta}$ is to predict the preferred sample $\bm{x}_{w}$, given the condition $\bm{c}$, as formalized in Eq.~\ref{eq:base}:
\begin{equation}
    \begin{aligned}
        &L_{\text{base}} = \mathbb{E}_{\bm{c}, \bm{x}_{w}} ||\bm{x}_{w} - f_{\theta}(\bm{c})||_2^2.
    \end{aligned}
    \label{eq:base_supp1}
\end{equation}

Minimizing the above Mean Square Error(MSE) is a well-established approach, equivalent to performing maximum likelihood estimation (MLE) in regression settings~\cite{1994estimating,learningreg,balancedmse}. Within this framework, $f_{\theta}(\bm{c})$ predicts the mean of a noisy distribution, which is assumed to follow a Gaussian distribution with constant variance $\sigma \bm{\text{I}}$, consistent with the probabilistic interpretation~\cite{glm}:
\begin{equation}
    \begin{aligned}
        p_{\theta} (\bm{x}_{w} | \bm{c}) = N(\bm{x}_{w} | f_{\theta}(\bm{c}), \sigma \bm{\text{I}}).
    \end{aligned}
    \label{eq:base_supp2}
\end{equation}

The MSE in Eq.~\ref{eq:base_supp1} equals the negative log-likelihood (NLL) of $p_{\theta} (\bm{x}_{w} | \bm{c})$~\cite{1994estimating}. Consequently, training the regression model $f_{\theta}$ using MSE implicitly enables it to approximate the conditional data distribution $p_{\text{data}}(\bm{x}_{w} | \bm{c})$.

\subsection{Preference Objective}

Besides the basics, we also draw inspiration from direct preference optimization (DPO)~\cite{dpo} and self-play finetuning (SPIN)~\cite{spin} to enhance alignment. These preference learning techniques adhere to a common RLHF principle~\cite{dpollm}: optimize the conditional distribution $p_{\theta} (\bm{x} | \bm{c})$ to maximize a latent reward model $r(\bm{c}, \bm{x})$, while regularizing the KL-divergence from a reference distribution $p_{\text{ref}}$:
\begin{equation}
    \begin{aligned}
    \underset{p_{\theta}}{\mathrm{max}}\; \mathbb{E}_{\bm{c},\bm{x}}[r(\bm{c}, \bm{x})] - 
    \eta \mathrm{KL} \;(p_{\theta}(\bm{x} | \bm{c})||p_{\text{ref}}(\bm{x} | \bm{c})).\label{eq:pref_supp1}
\end{aligned}
\end{equation}
Here $p_{\theta}$ and $p_{\text{ref}}$ are prediction distributions of $f_{\theta}$ and $f_{\text{ref}}$, respectively, where $f_{\text{ref}}$ is a copy of $f_{\theta}$ from an earlier training iteration, as defined in Eq.~\ref{eq:base_supp2}. The hyperparameter $\eta$ controls the strength of the regularization.

As demonstrated in~\cite{dpollm}, the unique global optimal solution of $p_{\theta}(\bm{x} | \bm{c})$ in Eq.~\ref{eq:pref_supp1} is expressed as:
\begin{equation}
    \begin{aligned}
    p_{\theta}(\bm{x} | \bm{c}) = p_{\text{ref}}(\bm{x} | \bm{c}) \exp\left({r(\bm{c}, \bm{x})}/{\eta}\right) / Z(\bm{c}),
\end{aligned}
\end{equation}
where $Z(\bm{c}) = \sum_{\bm{x}_0} p_{\text{ref}}(\bm{x}_0 | \bm{c}) \exp\left({r(\bm{c}, \bm{x}_0)}/{\eta}\right)$ is the partition function. The reward model is reformulated as:
\begin{equation}
    \begin{aligned}
    r(\bm{c}, \bm{x}) = \eta \log \frac{p_{\theta}(\bm{x} |\bm{c})}{p_{\text{ref}}(\bm{x} | \bm{c})} + \eta \log Z(\bm{c}).
\end{aligned}\label{eq:reward_supp}
\end{equation}

From the perspective of integral probability metric (IPM)~\cite{ipm}, DPO~\cite{dpo} maximizes the reward gap between preferred and non-preferred data distributions, while SPIN~\cite{spin} maximizes the reward gap between preferred data distribution and reference data distribution, \ie, $\bm{x}_{\text{ref}}=f_{\text{ref}}(\bm{c})\sim p_{\text{ref}}(\bm{x} |\bm{c})$. As introduced in~\cref{sec:ob}, we establish a combined objective of DPO and SPIN:
\begin{equation}
    \begin{aligned}
        \underset{r}{\mathrm{max}}\; \mathrm{E}_{\bm{c},\bm{x}_{w}, \bm{x}_{l}, \bm{x}_{\text{ref}}} 
        &[\underbrace{r(\bm{c}, \bm{x}_{w}) - r(\bm{c}, \bm{x}_{l})}_{\text{DPO}} \\&+ \mu(\underbrace{r(\bm{c}, \bm{x}_{w}) - r(\bm{c}, \bm{x}_{\text{ref}})}_{\text{SPIN}})],
    \end{aligned}\label{eq:ob_supp1}
\end{equation}
where $\mu$ is a hyperparameter that controls the trade-off. As demonstrated by~\cite{spinllm}, a more general form of the optimization problem in Eq.~\ref{eq:ob_supp1} is:
\begin{equation}
    \begin{aligned}
        \underset{r}{\mathrm{min}}\; \mathrm{E}_{\bm{c},\bm{x}_{w}, \bm{x}_{l}, \bm{x}_{\text{ref}}} 
        &[\ell(r(\bm{c}, \bm{x}_{w}) - r(\bm{c}, \bm{x}_{l}) \\&+ \mu(r(\bm{c}, \bm{x}_{w}) - r(\bm{c}, \bm{x}_{\text{ref}})))],
    \end{aligned}\label{eq:ob_supp2}
\end{equation}
where $\ell$ represents any monotonically decreasing convex loss function. Eq.~\ref{eq:ob_supp1} can be viewed as the maximization version of Eq.~\ref{eq:ob_supp2}, where $l(a) = -a$. However, using such a linear loss function leads to an unbounded objective value, which may cause undesirable negative infinite values of $r(\bm{c}, \bm{x}_{l})$ and $r(\bm{c}, \bm{x}_{\text{ref}})$ during continuous training. To address this issue, we adopt a logistic loss function as suggested by~\cite{dpo,spin}:
\begin{equation}
    l(a) := -\log \text{sigmoid} (a) = \log(1+\exp(-a)),
    \label{eq:ell}
\end{equation}
which is non-negative, smooth, and exhibits an exponentially decaying tail as $a\rightarrow \infty$. The logistic loss function helps prevent the excessive growth of the reward value $r$, ensuring a stable training process.

By substituting the reward model $r$ in Eq.~\ref{eq:ob_supp2} with Eq.~\ref{eq:reward_supp} and empirically setting $\eta$ and $\mu$ to 1, we obtain the final preference objective as follows:
\begin{equation}
    \begin{aligned}
L_{\text{pref}} &= \mathbb{E}_{\bm{c}, \bm{x}_w, \bm{x}_l, \bm{x}_{\text{ref}}} \left[ \ell \left( \log \frac{p_{\theta}(\bm{x}_w \vert \mathbf{c})}{p_{\text{ref}}(\bm{x}_w \vert \mathbf{c})} - \log \frac{p_{\theta}(\bm{x}_l \vert \mathbf{c})}{p_{\text{ref}}(\bm{x}_{l} \vert \mathbf{c})}  \right.\right.\\
&\left.\left. +  \log \frac{p_{\theta}(\bm{x}_{w} \vert \mathbf{c})}{p_{\text{ref}}(\bm{x}_{w} \vert \mathbf{c})} - \log \frac{p_{\theta}(\bm{x}_{\text{ref}} \vert \mathbf{c})}{p_{\text{ref}}(\bm{x}_{\text{ref}} \vert \mathbf{c})} \right) \right],\\
    \end{aligned}
    \label{eq:pref_supp3}
\end{equation}
which aligns with Eq.~\ref{eq:pref1}. Using the equivalence between MSE and NLL under the Gaussian prior, as discussed in~\cref{sec:bo}, we obtain a simplified version of $L_{\text{pref}}$ for implementation as follows:
\begin{equation}
    \begin{aligned}
L_{\text{pref}} &= \mathbb{E}_{\bm{c}, \bm{x}_{w}, \bm{x}_{l}} [\ell(-[2(||\bm{x}_{w} - f_{\theta}(\bm{c})||_2^2 - ||\bm{x}_{w} - f_{\text{ref}}(\bm{c})||_2^2) \\
&- (||\bm{x}_{l} - f_{\theta}(\bm{c})||_2^2 - ||\bm{x}_{l} - f_{\text{ref}}(\bm{c})||_2^2)\\ &- ||f_{\text{ref}}(\bm{c}) - f_{\theta}(\bm{c})||_2^2])],
    \end{aligned}
    \label{eq:pref_supp2}
\end{equation}
which is consistent with Eq.~\ref{eq:pref2}. As discussed in~\cref{sec:ob}, the reference model $f_{\text{ref}}$ is iteratively updated. Specifically, we first randomly initialize $f_{\text{ref}}$ and later iteratively copy $f_{\theta}$ to $f_{\text{ref}}$ whenever $f_{\theta}$ outperforms $f_{\text{ref}}$.
In practice, we execute the substitution when $f_{\theta}(c)$ is closer to $\bm{x}_{w}$ than $f_{\text{ref}}(c)$ for $k$ consecutive iterations, \ie, 
\begin{equation}
    \begin{aligned}
        ||\bm{x}_{w} - f_{\theta}(\bm{c})||_2^2 < ||\bm{x}_{w} - f_{\text{ref}}(\bm{c})||_2^2.
    \end{aligned}
\end{equation}

To summarize, The final Iteratively Updated Preference Objective is a combination of $L_{\text{base}}$ and $L_{\text{pref}}$, weighted by a ratio parameter $\lambda$:
\begin{equation}
    \begin{aligned}
    L = L_{\text{base}} + \lambda L_{\text{pref}}.
    \end{aligned}
    \label{eq:sum_supp}
\end{equation}

\section{Additional Quantitative Results}

In~\cref{tab:geneval}, we present additional quantitative results on GenEval~\cite{geneval} and DPGBench~\cite{ella}. IMG shows consistent improvements across two benchmarks.

\begin{table}[h]
    \centering
    \setlength{\tabcolsep}{2pt}
    \small
    \resizebox{0.7\linewidth}{!}{
    \begin{tabular}{lcc}
        \toprule
        \textbf{Model} & \textbf{GenEval↑} & \textbf{DPGBench↑} \\
        \midrule 
        SDXL-DPO & 0.59 & 76.81\\
        \rowcolor{gray!20} SDXL-DPO + IMG (Ours) & \textbf{0.61} & \textbf{78.72}\\
        \midrule 
        FLUX  & 0.68 & 80.60\\
        \rowcolor{gray!20} FLUX + IMG (Ours) & \textbf{0.70} & \textbf{82.77}\\
        \bottomrule
    \end{tabular}}
    \vspace{-2mm}
    \caption{Results on GenEval~\cite{geneval} and DPGBench~\cite{ella}.}
    \label{tab:geneval}
    \vspace{-2mm}
\end{table}

\section{Additional Qualitative Results}

\begin{figure}[h]
    \centering
    \vspace{-2mm}  
    \includegraphics[width=0.95
    \linewidth]{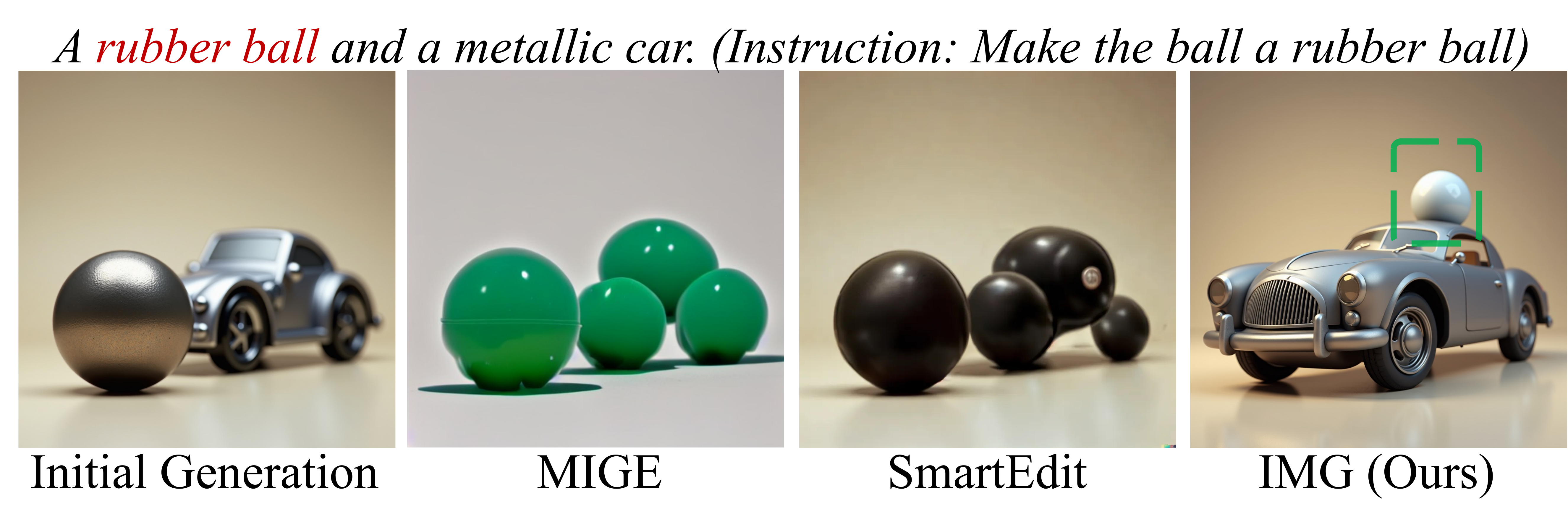}
    \vspace{-3mm}   
    \caption{Comparison between MLLM-based editing and IMG.}
    \label{fig:suppedit}
    \vspace{-5mm}
\end{figure}

In~\cref{fig:suppedit}, we compare IMG with leading MLLM-based image editing methods~\cite{MIGE,smartedit}. IMG showcases better alignment performance and visual quality.

\begin{figure*}[t]
    \centering
    \includegraphics[width=\linewidth]{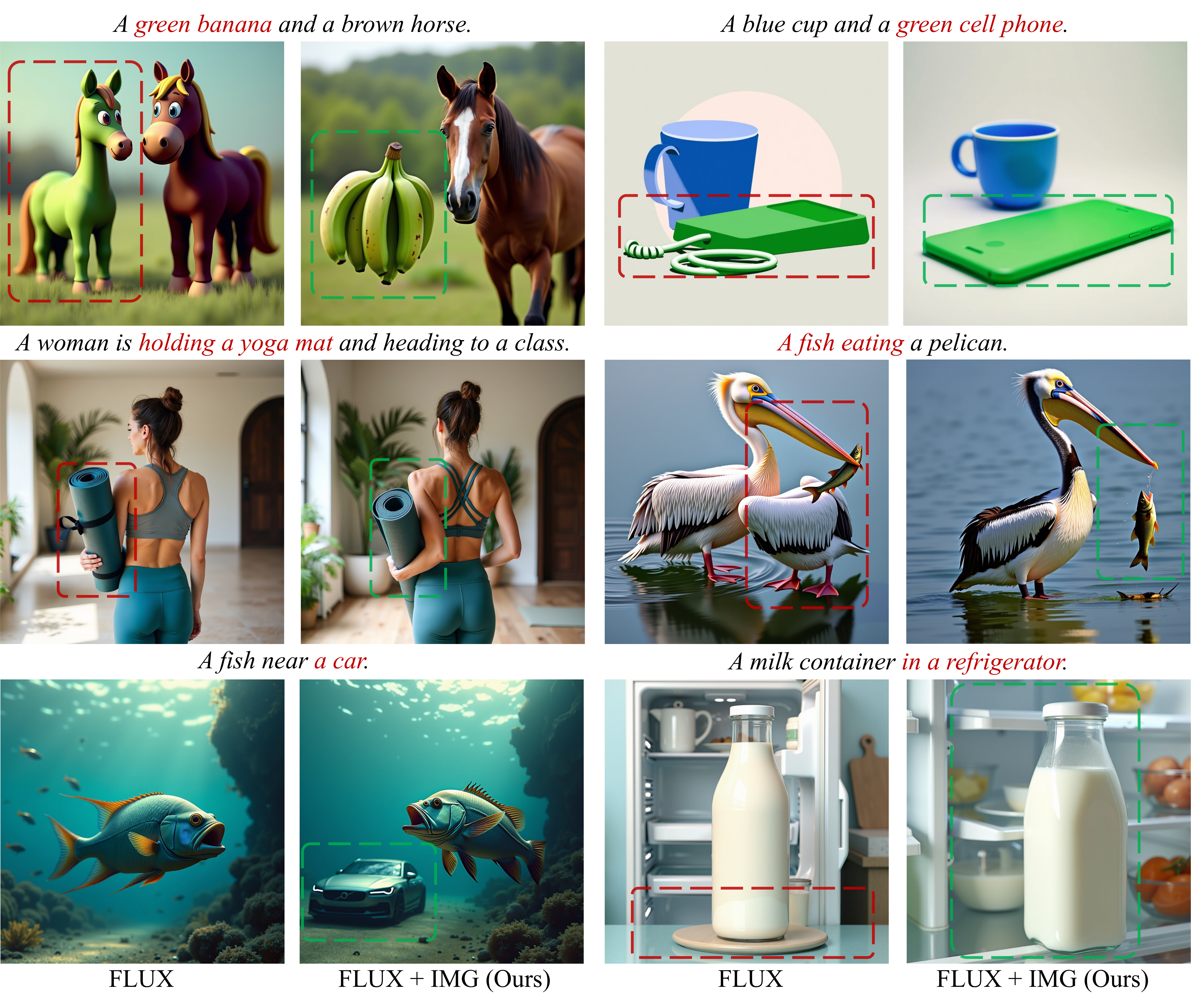}
      \vspace{-4mm}   
    \caption{{Additional qualitative results by integrating IMG with FLUX.}}
    \label{fig:supp1}\vspace{-2mm}
\end{figure*}

\begin{figure*}[t]
    \centering
    \includegraphics[width=\linewidth]{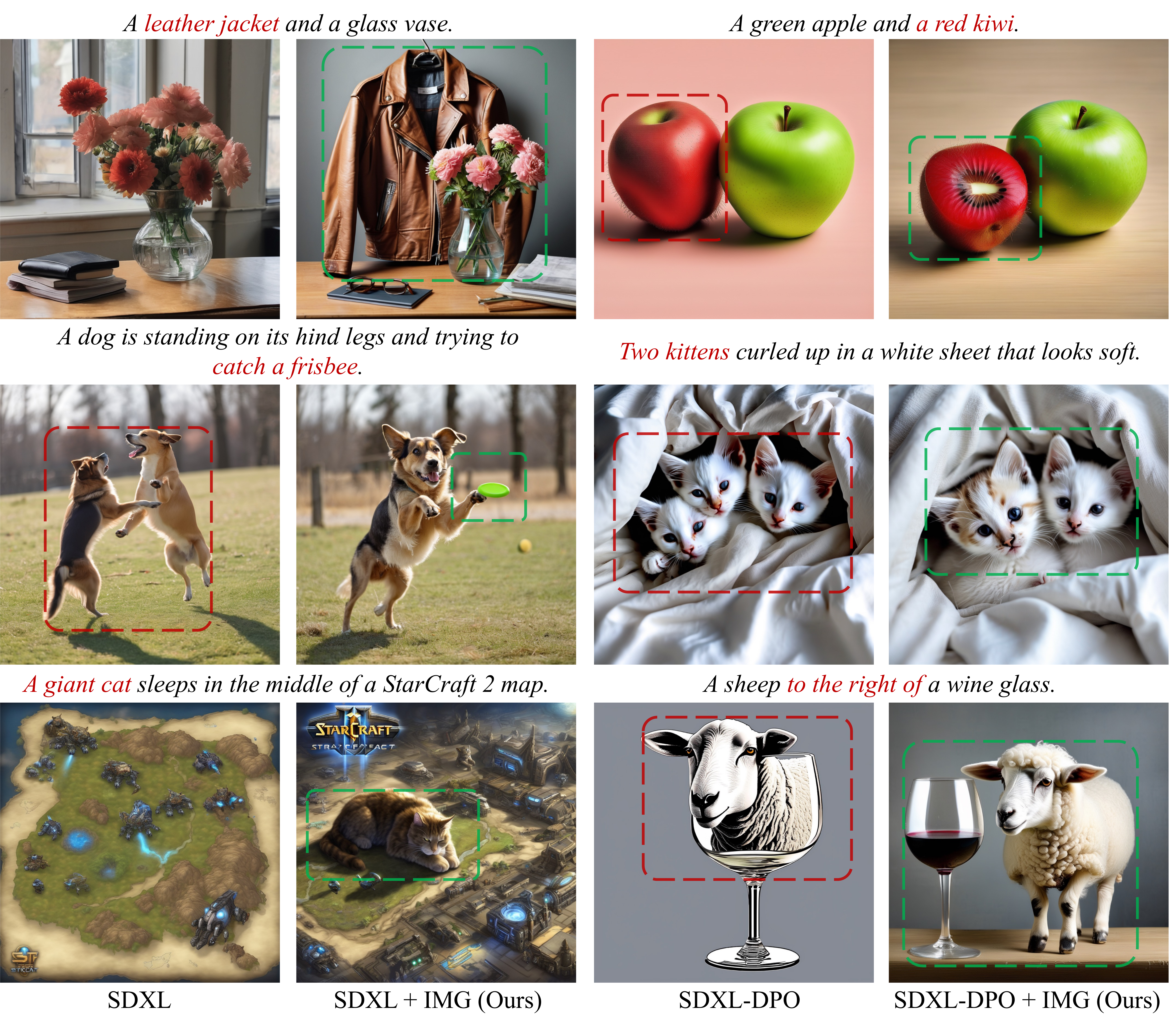}
      \vspace{-4mm}   
    \caption{{Additional qualitative results by integrating IMG with SDXL and SDXL-DPO.}}
    \label{fig:supp2}\vspace{-2mm}
\end{figure*}

In~\cref{fig:supp1} and~\cref{fig:supp2}, we present additional qualitative results to show the superior prompt adherence and aesthetic quality achieved by integrating IMG with various models.

